\documentclass[sigconf]{acmart}

\usepackage{balance}
  
\usepackage{subfigure}
\usepackage{tikz}

\def\BibTeX{{\rm B\kern-.05em{\sc i\kern-.025em b}\kern-.08emT\kern-.1667em\lower.7ex\hbox{E}\kern-.125emX}}
    
\copyrightyear{2020} 
\acmYear{2020} 
\setcopyright{rightsretained} 
\acmConference[PODS'20]{Proceedings of the 39th ACM SIGMOD-SIGACT-SIGAI Symposium on Principles of Database Systems}{June 14--19, 2020}{Portland, OR, USA}
\acmBooktitle{Proceedings of the 39th ACM SIGMOD-SIGACT-SIGAI Symposium on Principles of Database Systems (PODS'20), June 14--19, 2020, Portland, OR, USA}
\acmDOI{10.1145/3375395.3389131}
\acmISBN{978-1-4503-7108-7/20/06}

\def\NP{\mathsf{NP}}
\def\PP{\mathsf{PP}}
\def\NPPP{\NP^{\PP}}
\def\PPPP{\PP^{\PP}}
\def\coNP{co{\NP}}
\def\sp{\#\mathsf{P}}

\def\sat{{\sc Sat}}
\def\ms{{\sc MajSat}}
\def\ems{{\sc E-MajSat}}
\def\mms{{\sc MajMajSat}}
\def\ssat{\#{\sc Sat}}
\def\wmc{{\sc WMC}}

\def\dmpe{{\sc D-MPE}}
\def\dmar{{\sc D-MAR}}
\def\dmap{{\sc D-MAP}}
\def\dsdp{{\sc D-SDP}}

\def\mpe{{\sc MPE}}
\def\mar{{\sc MAR}}
\def\map{{\sc MAP}}
\def\sdp{{\sc SDP}}

\def\pr{{\it Pr}}

\def\X{{\bf X}}
\def\x{{\bf x}}
\def\Y{{\bf Y}}
\def\y{{\bf y}}
\def\Z{{\bf Z}}
\def\z{{\bf z}}

\def\sex{{\sc sex}}
\def\cond{{\sc c}}
\def\tea{{\sc t1}}
\def\teb{{\sc t2}}
\def\agree{{\sc agree}}

\def\pos{{+ve}}

\def\ace{{\sc ace}}
\def\ctd{{\sc c2d}}
\def\mctd{{\sc mini-c2d}}
\def\df{{\sc d4}}
\def\cache{{\sc cache}}
\def\dsharp{{\sc dsharp}}
\def\sharps{{\sc sharp-sat}}
\def\sdd{{\sc sdd}}
\def\cudd{{\sc cudd}}

\newcommand\shrink[1]{}
\newcommand\spara[1]{\vspace{2mm}\noindent {\bf #1}}

\begin{document}

\fancyhead{}

\title{Three Modern Roles for Logic in AI}

\author{Adnan Darwiche}
\affiliation{%
  \institution{Computer Science Department}
  \streetaddress{University of California, Los Angeles}
  \city{University of California}
  \state{Los Angeles}
}
\email{darwiche@cs.ucla.edu}

%
\renewcommand{\shortauthors}{Adnan Darwiche}

%
\begin{abstract}
We consider three modern roles for logic in artificial intelligence, which are based on the theory of tractable Boolean circuits:
(1)~logic as a basis for computation,
(2)~logic for learning from a combination of data and knowledge, and 
(3)~logic for reasoning about the behavior of machine learning systems.
\end{abstract}

%
%

\ccsdesc[500]{Computing methodologies~Learning in probabilistic graphical models}
\ccsdesc[500]{Computing methodologies~Logical and relational learning}
\ccsdesc[500]{Theory of computation~Automated reasoning}
\ccsdesc[500]{Theory of computation~Complexity classes}
\ccsdesc[500]{Theory of computation~Problems, reductions and completeness}

%
\keywords{tractable circuits, knowledge compilation, explainable AI}

%
\begin{teaserfigure}
 \includegraphics[width=\textwidth]{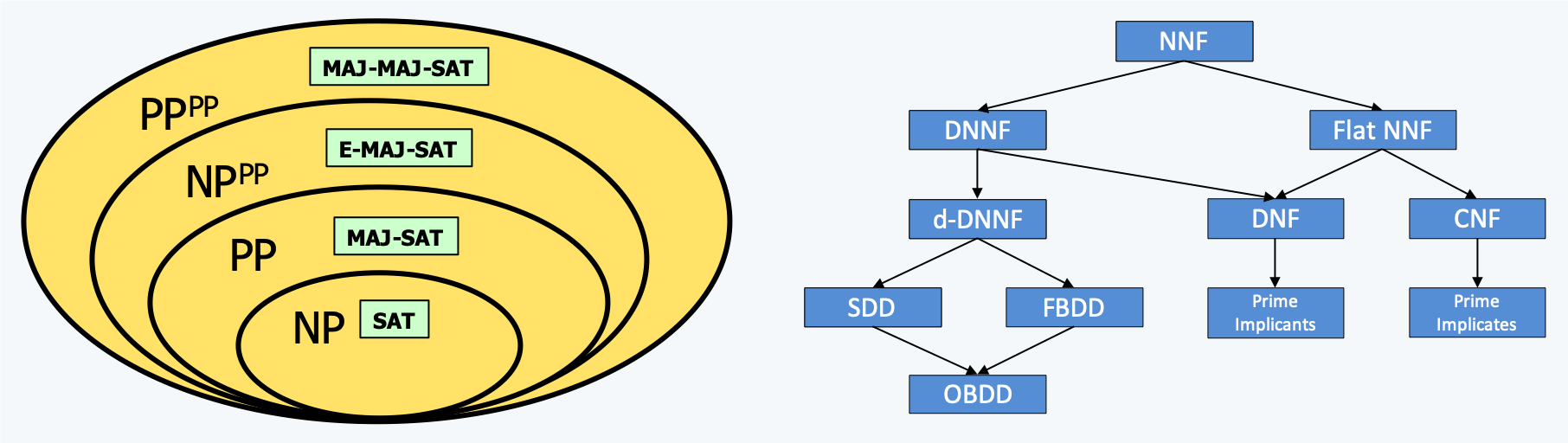}
  \caption{Tractable Boolean circuits as a basis for computation.  
  \label{fig:teaser}}
  \Description{Tractable Boolean circuits as a basis for computation.}
\end{teaserfigure}

%
\maketitle

\section{Introduction}

Logic has played a fundamental role in artificial intelligence since the field was incepted~\cite{McCarthy59}.
This  role has been mostly in the area of knowledge representation and reasoning, where logic is 
used to represent categorical knowledge and then draw conclusions based on deduction and other more 
advanced forms of reasoning. Starting with~\cite{nilsson1986probabilistic}, logic also formed the basis for 
drawing conclusions from a mixture of categorial and probabilistic knowledge.

In this paper, we review three modern roles for propositional logic in artificial intelligence, which are based 
on the theory of tractable Boolean circuits. This theory, which matured considerably during the last two decades, 
is based on Boolean circuits in Negation Normal Form (NNF) form. NNF circuits are not 
tractable, but they become tractable once we impose certain properties on them~\cite{darwicheJAIR02}.
Over the last two decades, this class of circuits has been studied systematically across three
dimensions. The first dimension concerns a synthesis of NNF circuits that have varying degrees
of tractability (the polytime queries they support). The second dimension concerns the relative
succinctness of different classes of tractable NNF circuits (the optimal size circuits can attain).
The third dimension concerns the development of algorithms for compiling Boolean formula into
tractable NNF circuits. 

The first modern role for logic we consider is in using tractable circuits as a basis for
computation, where we show how problems in the complexity classes 
$\NP$, $\PP$, $\NPPP$ and $\PPPP$ can be solved by compiling Boolean formula into corresponding
tractable circuits. These are rich complexity classes, which include some commonly utilized
problems from probabilistic reasoning and machine learning. We discuss this first role in two
steps. In Section~\ref{sec:computation}, we discuss the prototypical problems that
are complete for these complexity classes, which are all problems on Boolean formula. We also discuss
problems from probabilistic reasoning which are complete for these classes and their
reduction to prototypical problems. In Section~\ref{sec:circuits},
we introduce the theory of tractable circuits with exposure to circuit types that can be used
to efficiently solve problems in these complexity classes (if compiled successfully).

The second role for logic we shall consider is in learning from a combination of data and symbolic 
knowledge. We show again that this task can be reduced to a process of compiling, then reasoning
with, tractable circuits. This is discussed in Section~\ref{sec:bk}, where we also introduce and employ
a class of tractable probabilistic circuits that are based on tractable Boolean circuits. 

The third role for logic that we consider is in meta-reasoning, where we employ it to reason
about the behavior of machine learning systems. In this role, some common machine learning
classifiers are compiled intro tractable circuits that have the same input-output behavior. 
Queries pertaining to the explainability and robustness of decisions can then be answered
efficiently, while also allowing one to formally prove global properties of the underlying machine
learning classifiers. This role is discussed in Section~\ref{sec:meta}.

The paper concludes in Section~\ref{sec:outlook} with a perspective on
research directions that can mostly benefit the roles we shall discuss,
and a perspective on how recent developments in AI have triggered some
transitions that remain widely unnoticed. 

\section{Logic For Computation}
\label{sec:computation}

\begin{figure*}[tb]
  \centering
  \includegraphics[width=.50\linewidth]{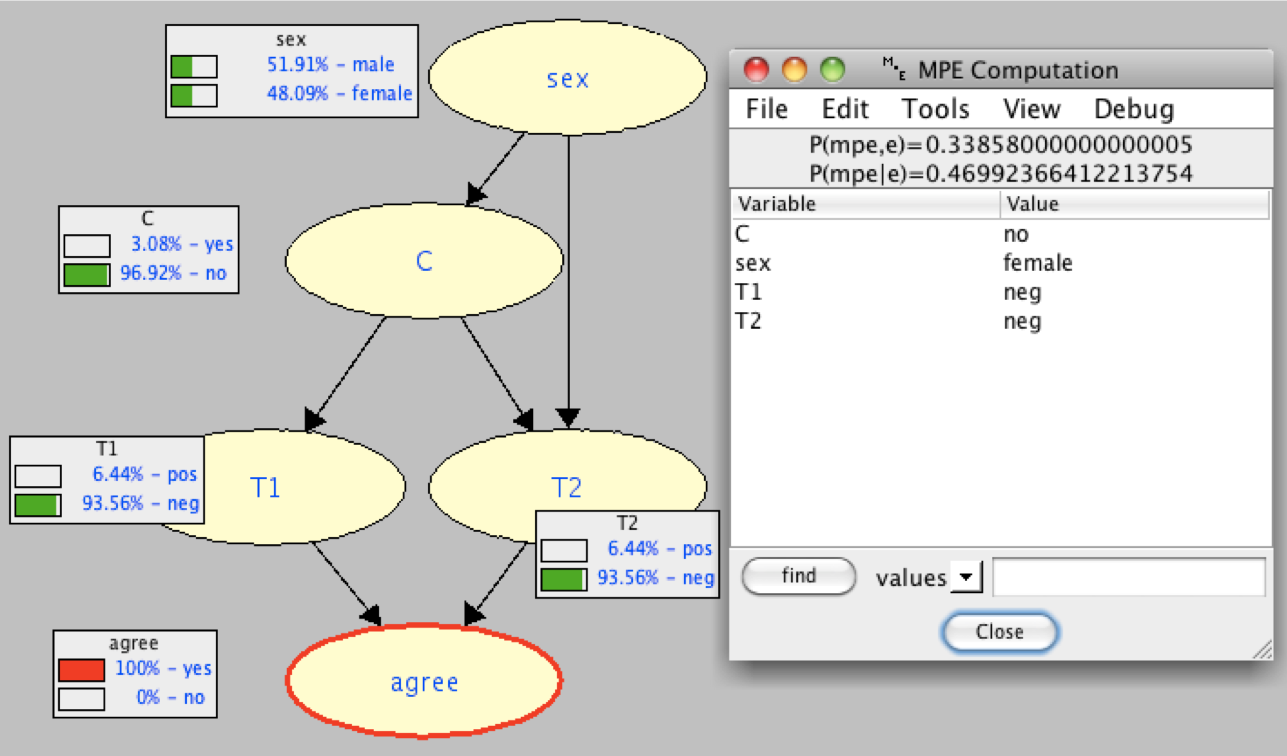}
  \hspace{5mm}
    \includegraphics[width=.35\linewidth]{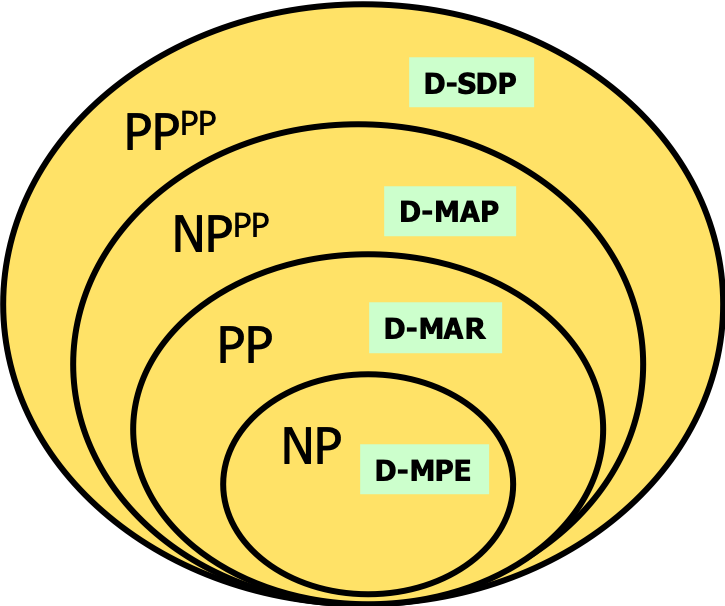}
  \caption{Left: A Bayesian network with an illustration of the \mpe\ query. The network concerns a
  medical condition \cond\ and two tests \tea\ and \teb\ that can be used to detect the condition.
  The variable \agree\ indicates whether the two test results are in agreement.
  Right: Decision problems on Bayesian networks that are complete for the classes 
  $\NP$, $\PP$, $\NPPP$ and $\PPPP$.
  \label{fig:bn-queries} \label{fig:bn-classes}}
    \Description{Complexity of Bayesian network queries.}
\end{figure*}

The first role we shall consider for logic is that of systematically solving problems in the complexity classes
$\NP$, $\PP$, $\NPPP$ and $\PPPP$, which are related in the following way:
$$
\NP \subseteq \PP \subseteq \NPPP \subseteq \PPPP.
$$
The prototypical problems for these complexity classes all correspond to questions on Boolean formula.
Moreover, these classes include problems that are commonly used in the areas of probabilistic
reasoning and machine learning. While there is a long tradition of developing dedicated 
algorithms when tackling problems in these complexity classes, it is now common to
solve such problems by reducing them to their Boolean counterparts, especially for
problems with no tradition of dedicated algorithms. 

Let us first consider four common problems from probabilistic reasoning
and machine learning that are complete for the classes $\NP$, $\PP$, $\NPPP$ and $\PPPP$.
These problems can all be stated on probability distributions specified using
Bayesian networks \cite{Darwiche09,Pearl88b,KollerFriedman,MurphyBook}.
These networks are directed acyclic graphs with nodes representing (discrete) variables.
Figure~\ref{fig:bn-queries} depicts a Bayesian network with five variables \sex, \cond, \tea,
\teb\ and \agree. 
The structure of a Bayesian network encodes conditional independence relations among its variables. 
Each network variable is associated with a set of distributions that are conditioned on the states
of its parents (not shown in Figure~\ref{fig:bn-queries}). 
These conditional probabilities and the 
conditional independences encoded by the network structure are satisfied by
exactly one probability distribution over the network variables. There are four question
on this distribution whose decision versions are complete for the complexity classes
$\NP$, $\PP$, $\NPPP$ and $\PPPP$.
Before we discuss these problems, we settle some notation first.

Upper case letters (e.g., $X$) will denote variables and lower case letters
(e.g., $x$) will denote their instantiations. That is, \(x\) is a {\em literal} specifying
a value for variable \(X\). Bold upper case letters (e.g., $\X$) will denote
sets of variables and bold lower case letters (e.g., $\x$) will denote their
instantiations. We liberally treat an instantiation of a variable set as
conjunction of its corresponding literals. 
\shrink{Given instantiations $\x$ and $\y$, we say $\x$ is {\em compatible} 
with $\y$, denoted $\x\sim\y$, iff $\x\wedge\y$ is satisfiable.}

The four problems we shall consider on a Bayesian network with variables \(\X\) 
and distribution \(\pr(\X)\) are \mpe, \mar, \map\ and \sdp.
The \mpe\ problem finds an instantiation \(\x\) of the network variables that has a maximal
probability \(\pr(\x)\). This is depicted in Figure~\ref{fig:bn-queries} which illustrates 
the result of an \mpe\ computation. 
The decision version of this problem, \dmpe, asks
whether there is a variable instantiation whose probability is greater than a given \(k\).
The decision problem \dmpe\ is complete for the class $\NP$~\cite{Shimony94}. 

We next have \mar\ which computes the probability of some value \(x\) for a variable \(X\). 
Figure~\ref{fig:bn-queries} depicts these probabilities for each variable/value pair.
The decision version, \dmar, asks whether \(\pr(x)\) is greater than a given \(k\).
\dmar\ is complete for the class $\PP$~\cite{Roth96}. \mar\ is perhaps the most
commonly used query on Bayesian networks and similar probabilistic graphical
models.

The next two problems are stated with respect to a subset of network variables 
\(\Y \subseteq \X\). The problem \map\ finds an instantiation \(\y\) that has
a maximal probability. For example, we may wish to find a most probable instantiation
of variables \sex\ and \cond\ in Figure~\ref{fig:bn-queries}. The decision version
of this problem, \dmap, asks whether there is
an instantiation \(\y\) whose probability is greater than a given \(k\). \dmap\
is complete for the class  $\NPPP$~\cite{ParkD04}.\footnote{Some treatments
in the literature use \map\ and {\em partial} \map\ when referring to \mpe\  and \map,
respectively.
Our treatment follows the original terminology used in~\cite{Pearl88b}.}

Suppose now that we are making a decision based on whether \(\pr(x) \geq T\) for some variable value
\(x\) and threshold \(T\). The \sdp\ problem finds the probability that 
this decision will stick after having observed the state of variables \(\Y\).
For example, we may want to operate on a patient if the probability of 
condition \cond\ in Figure~\ref{fig:bn-queries} is greater than \(90\%\) (the decision
is currently negative). The \sdp\ (same-decision probability) can be used to compute
the probability that this decision will stick after having obtained the results of tests \tea\ and \teb~\cite{DarwicheChoi10}.
The \sdp\ computes an expectation. Its decision version, \dsdp, asks whether this probability is greater than
a given \(k\) and is complete for the class $\PPPP$~\cite{ChoiXueDarwiche11}.
This query was used to assess the value of information with applications to reasoning about
features in Bayesian network classifiers; see, 
e.g.,~\cite{DBLP:conf/aaai/ChenCD15,DBLP:conf/ijcai/ChoiDB17,DBLP:conf/ijcai/ChoiB18}.

There is a tradition of solving the above problems using dedicated 
algorithms; see, e.g., the treatments in~\cite{Darwiche09,Pearl88b,KollerFriedman,MurphyBook}.
Today these problems are also being commonly solved using reductions to 
the prototypical problems of the corresponding complexity classes, which are 
defined over Boolean formula as mentioned earlier. This is particularly the
case for the \mar\ problem and for distributions that
are specified by representations that go beyond Bayesian networks;
see, e.g., \cite{NIPS2018_7632,DBLP:conf/cp/LatourBDKBN17,DBLP:conf/pkdd/DriesKMRBVR15}. 
These reduction-based approaches are the state of the art on certain problems; for example, 
when the Bayesian network has an abundance of 
0/1 probabilities or context-specific independence~\cite{UAIEvaluation08}.\footnote{Context-specific
independence refers to independence relations among network variables
which are implied by the specific probabilities that quantify the network
and, hence, cannot be detected by only examining the 
network structure~\cite{BoutilierFGK96}.}

We next discuss the prototypical problems of complexity classes 
$\NP$, $\PP$, $\NPPP$ and $\PPPP$ before we illustrate in Section~\ref{sec:reduction}
the core technique used in reductions to these problems. We then follow
in Section~\ref{sec:circuits} by showing how these prototypical problems can be solved
systematically by compiling Boolean formula into Boolean circuits with varying
degrees of tractability. 

\subsection{Prototypical Problems}

\begin{figure}[tb]
  \centering
  \includegraphics[width=\linewidth]{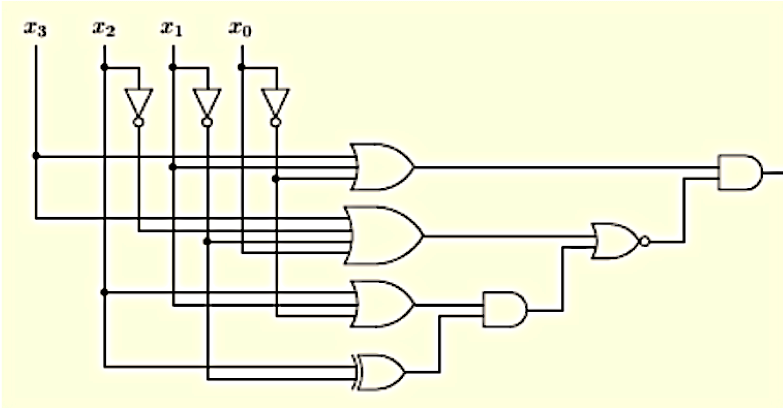}
  \caption{A Boolean circuit to illustrate the problems
  \sat, \ms, \ems\ and \mms.  \label{fig:circuit}}
  \Description{Sat, MajSat, E-MajSat, MajMajSat.}
\end{figure}

$\NP$ is the class of decision problems that can be solved by a
non-deterministic polynomial-time Turing machine. $\PP$ is the class of decision
problems that can be solved by a non-deterministic polynomial-time Turing
machine, which has more accepting than rejecting paths. $\NPPP$ and $\PPPP$ are
the corresponding classes assuming the corresponding
Turing machine has access to a $\PP$ oracle.

The prototypical problems for these classes are relatively simple and are usually
defined on Boolean formula in Conjunctive Normal Form (CNF). We will however
illustrate them on Boolean circuits to make a better connection to the discussion
in Section~\ref{sec:circuits}.

Consider a Boolean circuit \(\Delta\) which has input variables \(\X\) and let \(\Delta(\x)\) be the 
output of this circuit given input \(\x\); see Figure~\ref{fig:circuit}. The
following decision problems are respectively complete and prototypical for the
complexity classes $\NP$, $\PP$, $\NPPP$ and $\PPPP$.

\spara{---\sat} asks if there is a circuit input $\x$ such that $\Delta(\x)=1$.  

\spara{---\ms} asks if the majority of circuit inputs $\x$ are such that $\Delta(\x)=1$. 

\spara{}The next two problems require that we partition the circuit input variables \(\X\) 
into two sets \(\Y\) and \(\Z\).

\spara{---\ems} asks if there is a circuit input $\y$ such that \(\Delta(\y,\z)=1\) for the majority of circuit inputs $\z$.

\spara{---\mms} asks if the majority of circuit inputs $\y$ are such that \(\Delta(\y,\z)=1\) 
for the majority of circuit inputs $\z$. 

There are two functional versions of \ms\ which have been receiving increased attention recently.
The first is \ssat\ which asks for the number of circuit inputs \(\x\) such that \(\Delta(\x)=1\).
Algorithms that solve this problem are known as {\em model counters.}\footnote{Some of the
popular or traditional model counters are \ctd~\cite{Darwiche04}, \mctd~\cite{OztokD18}, 
\df~\cite{DBLP:conf/ijcai/LagniezM17}, \cache~\cite{SangBK05}, \sharps~\cite{DBLP:conf/sat/Thurley06},
\sdd~\cite{ChoiDarwiche13} and \dsharp~\cite{DBLP:conf/ai/MuiseMBH12}. Many of these systems can also compute weighted model counts.} 
The more general functional version of \ms\ and the one typically used in practical reductions is called {\em weighted
model counting,} \wmc. In this variant, each circuit input \(x\) for variable \(X\) is given a weight \(W(x)\).
A circuit input \(\x = x_1, \ldots, x_n\) is then assigned the weight \(W(\x) = W(x_1) \ldots W(x_n)\).
Instead of counting the number of inputs \(\x\) such that \(\Delta(\x)=1\), weighted model
counting adds up the weights of such inputs. That is, \wmc\ computes \(\sum_\x W(\x)\) for all
circuit inputs \(\x\) such that \(\Delta(\x)=1\). 

\subsection{The Core Reduction}
\label{sec:reduction}

\begin{figure}[tb]
  \centering
  \includegraphics[width=\linewidth]{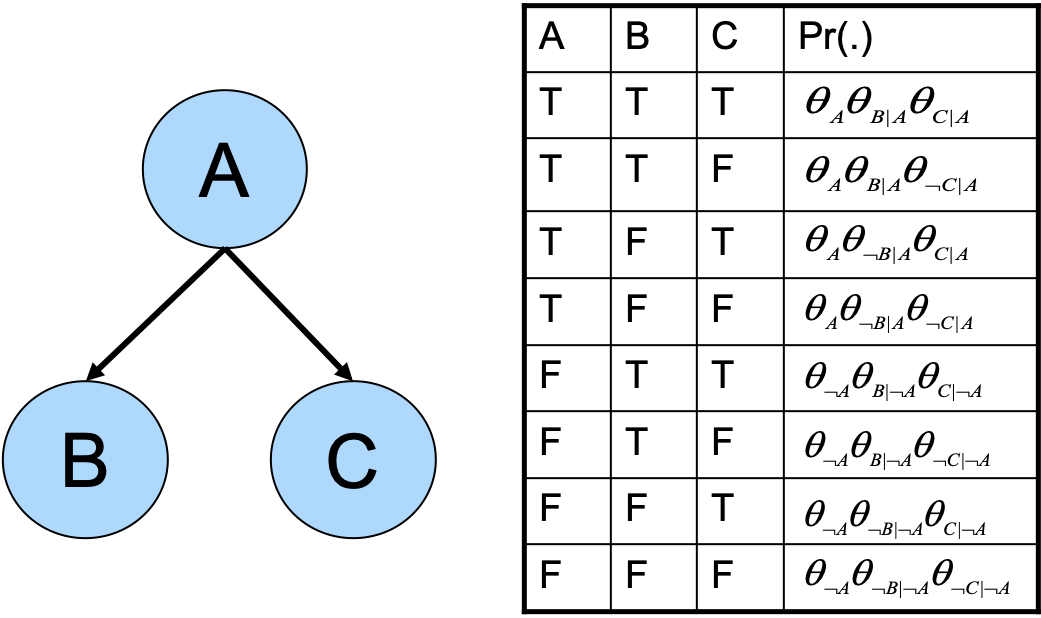}
  \caption{A Bayesian network and its distribution.  \label{fig:reduction}}
    \Description{The distribution of a Bayesian network.}
\end{figure}

As mentioned earlier, the decision problems \sat, \ms, \ems\ and \mms\ on Boolean formula are complete and prototypical for
the complexity classes $\NP$, $\PP$, $\NPPP$ and $\PPPP.$ The decision problems \dmpe, \dmar, \dmap\
and \dsdp\ on Bayesian networks are also complete for these complexity classes, respectively. Practical reductions of 
the latter problems to the former ones have been developed over the last two decades; see~\cite[Chapter 11]{Darwiche09} 
for a detailed treatment. Reductions have also been proposed from the functional problem \mar\ to \wmc, which are of 
most practical significance. We will next discuss the first such reduction~\cite{Darwiche02} since it is relatively 
simple yet gives the essence of how one can reduce problems that appear in probabilistic reasoning
and machine learning to problems on Boolean formula and circuits. In Section~\ref{sec:circuits}, we will further
show how these problems (of numeric nature) can be competitively solved using purely symbolic manipulations.

Consider the Bayesian network in Figure~\ref{fig:reduction}, which has three binary variables \(A\), \(B\) and \(C\).
Variable \(A\) has one distribution \((\theta_{A},\theta_{\neg A})\). Variable \(B\) has two distributions, which are
conditioned on the state of its parent \(A\): 
\((\theta_{B\vert A},\theta_{\neg B\vert A})\) and \((\theta_{B\vert \neg A},\theta_{\neg B\vert \neg A})\).
Variable \(C\) also has two similar distributions:
\((\theta_{C\vert A},\theta_{\neg C\vert A})\) and \((\theta_{C\vert \neg A},\theta_{\neg C\vert \neg A})\). 
We will refer to the probabilities \(\theta\) as {\em network parameters.} The Bayesian network in Figure~\ref{fig:reduction}
has ten parameters.

This Bayesian network induces the distribution depicted in Figure~\ref{fig:reduction}, where the probability
of each variable instantiation is simply the product of network parameters that are compatible with that
instantiation; see~\cite[Chapter 3]{Darwiche09} for a discussion of the syntax and semantics of Bayesian
networks. We will next show how one can efficiently construct a Boolean formula \(\Delta\) from a Bayesian network,
allowing one to compute marginal probabilities on the Bayesian network by performing weighted model
counting on formula \(\Delta\).

The main insight is to introduce a Boolean variable \(P\) for each network parameter \(\theta\),
which is meant to capture the presence or absence of parameter \(\theta\) given an instantiation
of the network variables (i.e., a row of the table in Figure~\ref{fig:reduction}). 
For the network in Figure~\ref{fig:reduction}, this leads to introducing ten Boolean variables: 
\(P_A\), \(P_{\neg A}\), \(P_{B\vert A}, \ldots, P_{\neg C \vert \neg A}\). In the second row of Figure~\ref{fig:reduction},
which corresponds to variable instantiation \(A, B, \neg C\), parameters \(\theta_A\), \(\theta_{B\vert A}\) and 
\(\theta_{\neg C\vert A}\) are present and the other seven parameters are absent. 

We can capture such presence/absence by adding one expression to the Boolean formula \(\Delta\) 
for each network parameter. For example, 
the parameters associated with variable \(A\) introduce the following expressions:
\(A \iff P_A\) and \(\neg A \iff \neg P_{\neg A}\). Similarly, the parameters of variable \(B\) introduce the
expressions \(A \wedge B \iff P_{B\vert A}\), \(\:\:A \wedge \neg B \iff P_{\neg B\vert A}\),
\(\:\:\neg A \wedge B \iff P_{B\vert \neg A}\) and \(\:\:\neg A \wedge \neg B \iff P_{\neg B\vert \neg A}\).
The parameters of variable \(C\) introduce similar expressions.

The resulting Boolean formula \(\Delta\) will have exactly eight models, which correspond to the 
network instantiations. The following is one of these models which correspond to instantiation
\(A, B, \neg C\):
\begin{eqnarray}
A \:\: B \:\: \neg C \:\: P_A \:\: P_{B\vert A} \:\: P_{\neg C\vert A} \nonumber \\ 
\neg P_{\neg A} \:\:\: \neg P_{\neg B\vert A} \neg P_{B\vert \neg A} \neg P_{\neg B\vert \neg A} \:\:\:
\neg P_{C\vert A} \neg P_{C\vert \neg A} \neg P_{\neg C\vert \neg A}.  \label{eq:exp}
\end{eqnarray}
In this model, all parameters associated with instantiation \(A, B, \neg C\) appear positively (present)
while others appear negatively (absent).

The last step is to assign weights to the values of variables (literals). For network variables, all
literals get a weight of \(1\); for example, \(W(A)=1\) and \(W(\neg A)=1\). 
The negative literals of parameter variables also get a weight of \(1\); for example, \(W(\neg P_A) = 1\)
and \(W(\neg P_{\neg C \vert A}) = 1\).
Finally, positive literals of network parameters get weights equal to these
parameters; for example, \(W(P_{A}) = \theta_{A}\) and \(W(P_{\neg C \vert A}) = \theta_{\neg C\vert A}\).
The weight of expression~(\ref{eq:exp}) is then \(\theta_{A}\theta_{B\vert A}\theta_{\neg C\vert A}\),
which is precisely the probability of network instantiation \(A, B, \neg C\). We can now compute
the probability of any Boolean expression \(\alpha\) by simply computing the weighted model
count of \(\Delta \wedge \alpha\), which completes the reduction of \mar\ to \wmc.

Another reduction was proposed in~\cite{SangBK05} which is suited towards Bayesian networks that
have variables with large cardinalities. More refined reductions have also been proposed which
can capture certain properties of network parameters such as 0/1 parameters and context-specific 
independence (can be critical for the efficient computation of weighted model counts on
the resulting Boolean formula). A  detailed treatment of reduction techniques and various practical
tradeoffs can be found in~\cite{Chavira.Darwiche.Aij.2008} 
and~\cite[Chapter 13]{Darwiche09}.\footnote{\ace\ implements some of these
reductions: \url{http://reasoning.cs.ucla.edu/ace/}}

\section{Tractable Circuits}
\label{sec:circuits}

\begin{figure}[tb]
  \centering
  \includegraphics[width=\linewidth]{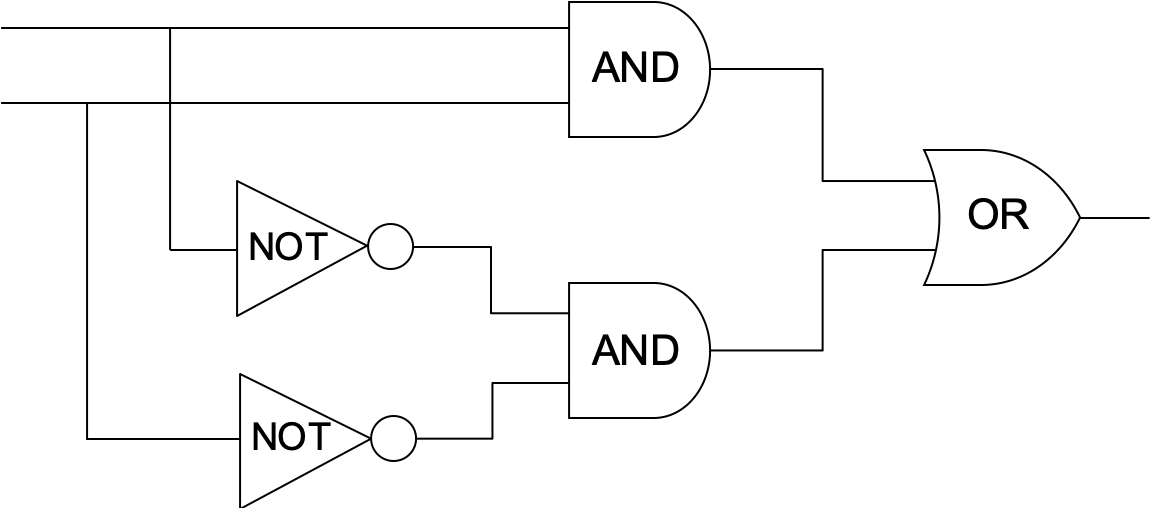}
  \caption{Negation Normal Form (NNF) circuit. \label{fig:nnf}}
    \Description{Negation Normal Form circuit (NNF circuit).}
\end{figure}

We now turn to a systematic approach for solving prototypical problems in the classes 
$\NP$, $\PP$, $\NPPP$ and $\PPPP,$ which is based on {\em compiling} Boolean formula
into tractable Boolean circuits. The circuits we shall compile into are in Negation Normal
Form (NNF) as depicted in Figure~\ref{fig:nnf}. These circuits have three types of gates:
and-gates, or-gates and inverters, except that inverters can only feed from the circuit
variables. Any circuit with these types of gates can be converted to an NNF circuit while
at most doubling its size.

NNF circuits are not tractable. However, by imposing certain properties on them we 
can attain different degrees of tractability. The results we shall review next are part
of the literature on {\em knowledge compilation,} an area that has been under 
development for a few decades, see, e.g.,~\cite{SelmanK96,CadoliD97,Marquis95}, except 
that it took a different turn since~\cite{darwicheJAIR02}; see also~\cite{Darwiche14}. 
Earlier work on knowledge compilation focused on {\em flat} NNF
circuits, which include subsets of Conjunctive Normal Form (CNF) and 
Disjunctive Normal Form (DNF) such as prime implicates, Horn clauses and prime implicants.
Later, however, the focus shifted towards {\em deep} NNF circuits with no
restriction on the number of circuit layers. A comprehensive treatment was 
initially given in~\cite{darwicheJAIR02}, in which some tractable NNF circuits were studied
across the two dimensions of {\em tractability} and {\em succinctness.} As we increase the 
strength of properties imposed on NNF circuits, their tractability increases by
 allowing more queries to be performed in polytime. This, however, typically
comes at the expense of succinctness as the size of circuits gets larger.

\begin{figure}[tb]
  \centering
  \includegraphics[width=\linewidth]{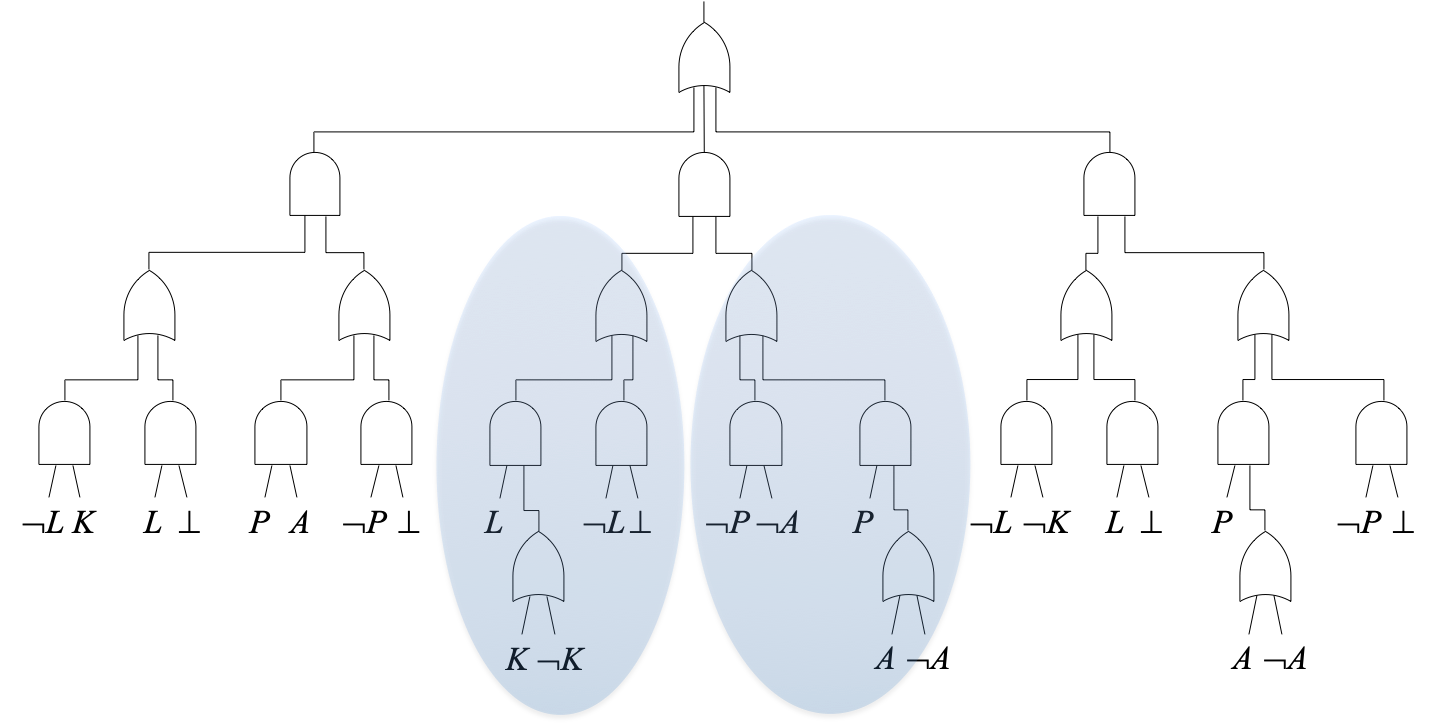}
  \caption{Illustrating the {\em decomposability} property of NNF circuits. 
The illustration does not tie shared inputs of the circuit for clarity of exposition. \label{fig:dnnf}}
  \Description{Decomposable NNF circuits (DNNF circuits).}
\end{figure}

One of the simplest properties that turn NNF circuits into tractable ones is
{\em decomposability}~\cite{darwicheJACM-DNNF}. According to this property, 
subcircuits feeding into an and-gate cannot share circuit variables. 
Figure~\ref{fig:dnnf} illustrates this property by highlighting the two 
subcircuits (in blue) feeding into an and-gate. The subcircuit on the left
feeds from circuit variables \(K\) and \(L\), while the one on the right
feeds from circuit variables \(A\) and \(P\). NNF circuits that satisfy the 
decomposability property are known as Decomposable NNF (DNNF) circuits.
The satisfiability of DNNF circuits can be decided in time linear in
the circuit size~\cite{darwicheJACM-DNNF}. Hence, enforcing decomposability
is sufficient to unlock the complexity class $\NP.$

\begin{figure}[tb]
\centering
\includegraphics[width=\linewidth]{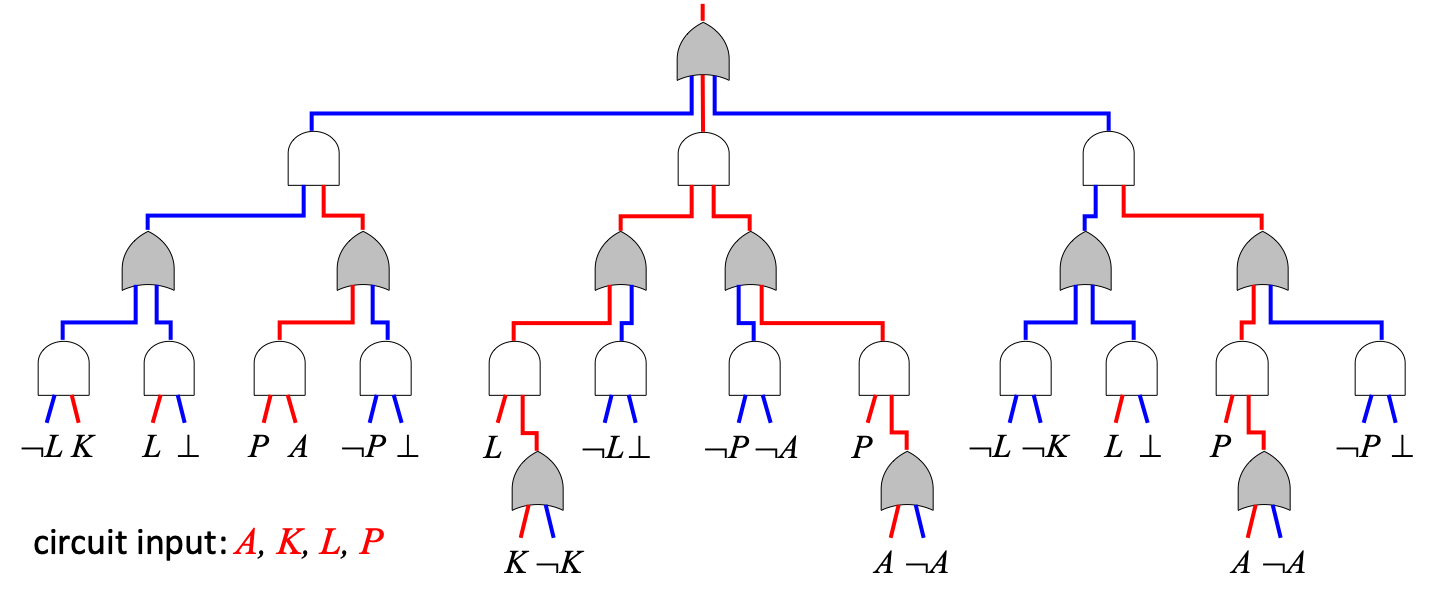}
\caption{Illustrating the {\em determinism} property of NNF circuits. Red wires are high and blue ones are low.
\label{fig:d-dnnf}}
  \Description{Deterministic, decomposable NNF circuits (d-DNNF circuits).}
\end{figure}

The next property we consider is {\em determinism}~\cite{darwiche01tractable}, which applies
to or-gates in an NNF circuit. According to this property, at most one input for an or-gate must
be high under any circuit input. Figure~\ref{fig:d-dnnf} illustrates this property when 
all circuit variables \(A, K, L, P\) are high. Examining the or-gates in this circuit, under this circuit input,
one sees that each or-gate has either one high input or no high inputs. This property corresponds to
mutual exclusiveness when the or-gate is viewed as a disjunction of its inputs.
\ms\ can be decided in polytime on NNF circuits that are both decomposable
and deterministic. These circuits are called d-DNNF circuits. If they are also
{\em smooth}~\cite{Darwiche03}, a property that can be enforced in quadratic time, 
d-DNNF circuits allow one to perform weighted model counting (\wmc) in linear time.\footnote{One
can actually compute all {\em marginal,} weighted model counts in linear time~\cite{darwiche01tractable}.}
The combination of decomposability and determinism therefore unlocks the complexity 
class $\PP$.

\begin{figure}[tb]
\centering
\includegraphics[width=\linewidth]{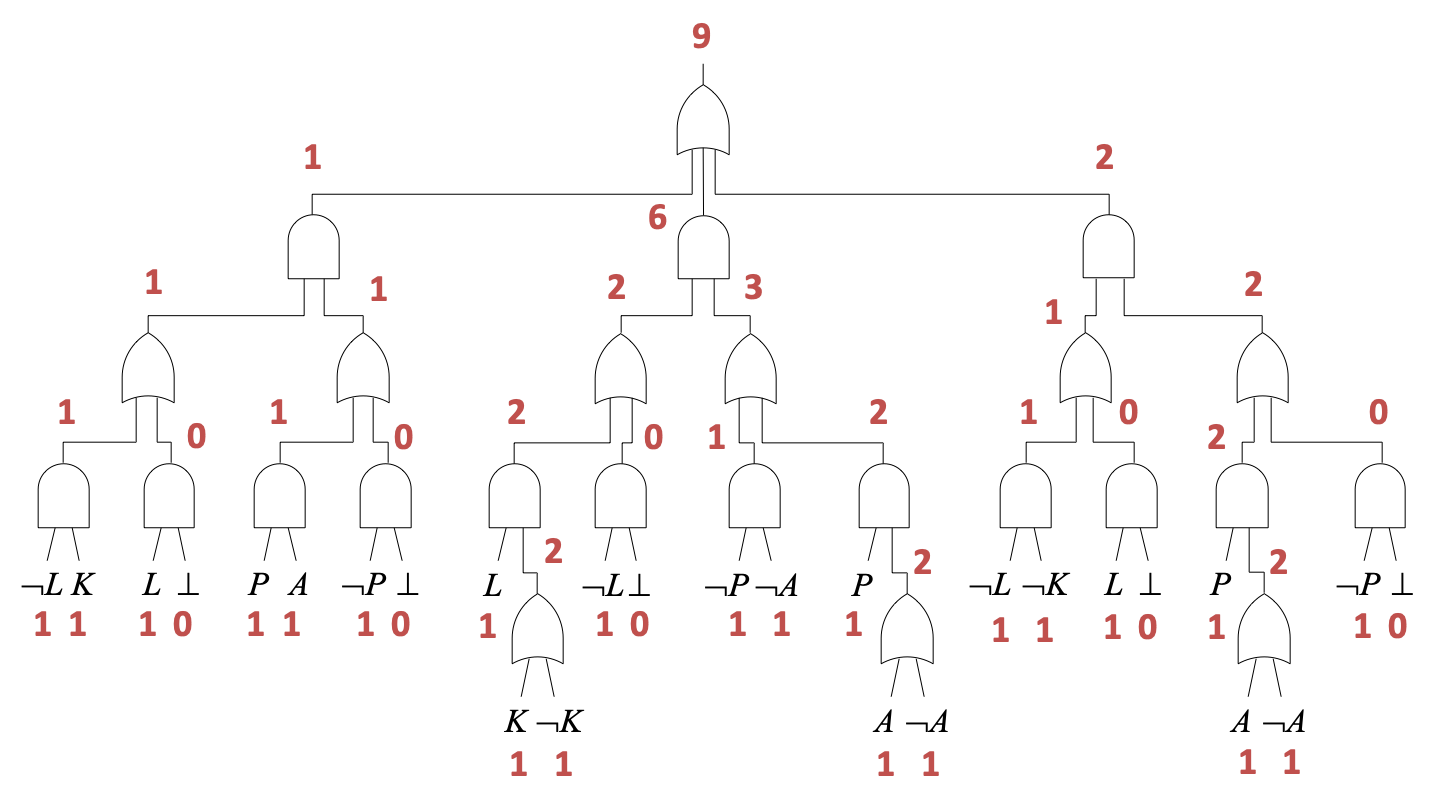}
\caption{Model counting in linear time on d-DNNF circuits. \label{fig:mc}}
  \Description{Model counting using d-DNNF circuits.}
\end{figure}

Smoothness requires that all subcircuits feeding into an or-gate mention the
same circuit variables. For example, in Figure~\ref{fig:d-dnnf}, three subcircuits 
feed into the top or-gate. Each of these subcircuits mentions the same set
of circuit variables:  \(A, K, L, P\). Enforcing smoothness can introduce trivial gates
into the circuit such as the bottom three or-gates in Figure~\ref{fig:d-dnnf} and
can sometimes be done quite efficiently~\cite{DBLP:conf/nips/ShihBBA19}.
An example of model counting using a d-DNNF circuit is depicted in
Figure~\ref{fig:mc}. Every circuit literal, whether a positive literal such as \(A\)
or a negative literal such as \(\neg A\), is assigned the value \(1\). Constant inputs
\(\top\) and \(\bot\) are assigned the values \(1\) and \(0\).
We then propagate these numbers upwards, multiplying numbers assigned to the 
inputs of an and-gate and summing numbers assigned to the inputs of an or-gate. 
The number we obtain for the circuit output is the model count. In this example, the circuit has \(9\) satisfying
inputs out of \(16\) possible ones. To perform weighted model counting,
we simply assign a weight to a literal instead of the value \(1\)---model counting (\ssat)
is a special case of weighted model counting (\wmc) when the weight of each literal is \(1\).

\begin{figure}[tb]
 \centering
\includegraphics[width=\linewidth]{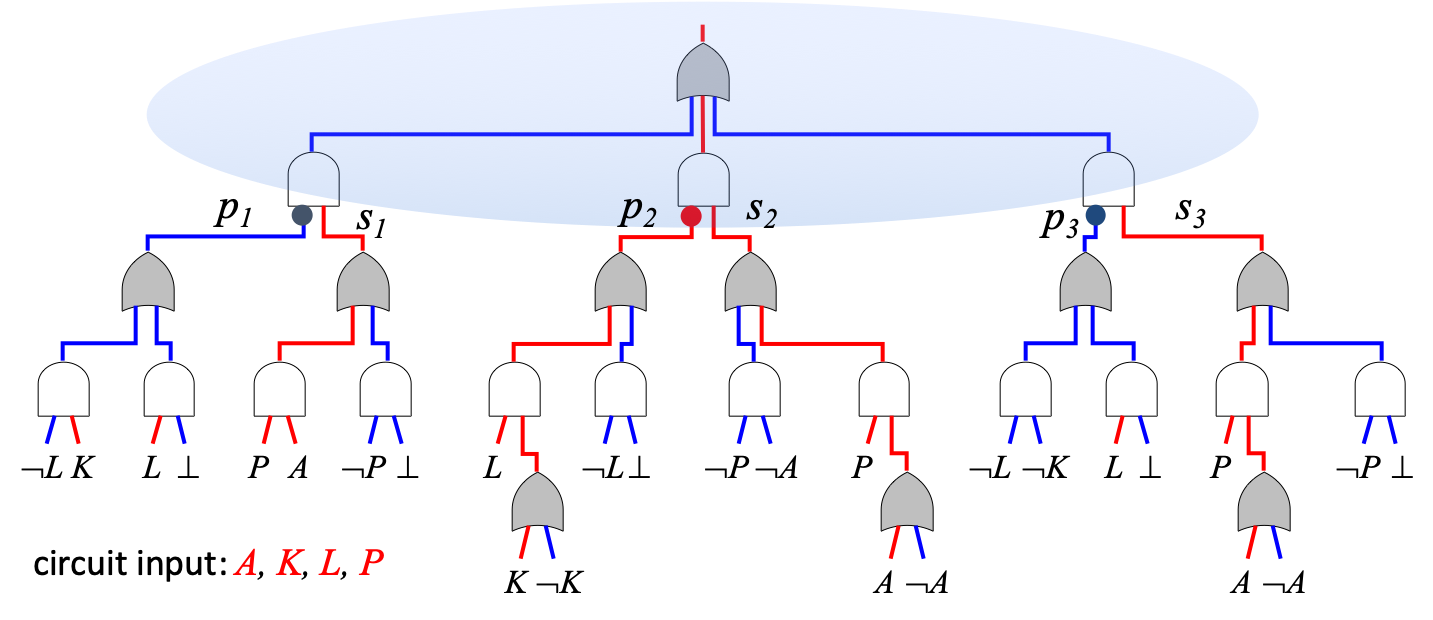}
\caption{Illustrating the {\em sentential decision} property of NNF circuits. Red wires are high and blue ones are low.
\label{fig:sdd}}
  \Description{Sentential Decision Diagrams (SDD circuits).}
\end{figure}

\begin{figure}[tb]
\centering
  \subfigure[vtree]{
    \label{fig:vtree}
\begin{tikzpicture}[-,
                   level 1/.style={sibling distance = .9cm, level distance = 1.1cm},
                   level 2/.style={sibling distance = .45cm, level distance = 1.1cm}] 
\node {\scriptsize 1}
  child{ 
    node {\scriptsize 2}
      child {node {$\scriptsize L$}}
      child {node {$\scriptsize K$}}
  }
  child{
    node {\scriptsize 3}
      child{node {$\scriptsize P$}}
      child{node {$\scriptsize A$}}
  };
\end{tikzpicture}
  }\hspace{5mm}
  \subfigure[constrained \(CE \vert ABD\)]{
  \label{fig:c-vtree}
 \begin{tikzpicture}[-,
                   level 1/.style={sibling distance = 1.4cm, level distance = .8cm},
                   level 2/.style={sibling distance = .6cm, level distance = .8cm}
                   ] 
\node {\scriptsize 1}
  child { 
    node {\scriptsize 2} 
      child {
        node {\scriptsize 3}
          child{node {$A$}}
          child{node { $B$}}
      }
      child{node { $D$}}
  }
  child {
    node {\scriptsize 4} 
    child{node { $C$}}
    child{node { $E$}}
  };
\end{tikzpicture}
  }\hspace{5mm}
    \subfigure[right-linear]{
    \label{fig:rl-vtree}
\begin{tikzpicture}[-,
                   level 1/.style={sibling distance = .7cm, level distance = 1.1cm},
                   level 2/.style={sibling distance = .7cm, level distance = 1.1cm}] 
\node {\scriptsize 1}
  child {node {$A$}}
  child{
    node {\scriptsize 2}
      child{node {$B$}}
      child{
      node {\scriptsize 3} 
     child{node {$C$}}
     child{node {$D$}}}
  };
\end{tikzpicture}
  }
\caption{A {\em constrained vtree} for \(\X\vert \Y\) is a vtree over variables \(\X \cup \Y\) that contains a 
node \(u\) with following properties: (1)~\(u\) can be reached from the vtree root by following right children only
and (2)~the variables of \(u\) are precisely \(\X\). A {\em right-linear} vtree is one in which the left child of every internal node is a leaf.
\label{fig:vtrees}}
  \Description{Variable trees (vtrees).}
\end{figure}
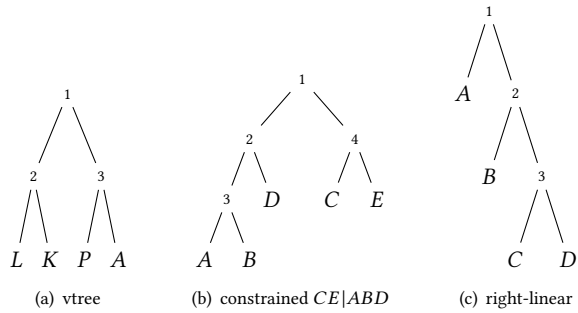

There are stronger versions of decomposability and determinism which give rise to additional,
tractable NNF circuits. {\em Structured decomposability} is stated with respect to a binary
tree whose leaves are in one-to-one correspondence with the circuit variables~\cite{PipatsrisawatD08}. 
Such a tree is depicted in Figure~\ref{fig:vtree} and is known as a {\em vtree.}
Structured decomposability requires that each and-gate has two inputs \(i_1\) and \(i_2\) 
and correspond to a vtree node \(v\) such that the variables of subcircuits feeding into \(i_1\)
and \(i_2\) are in the left and right subtrees of vtree node \(v\). The DNNF 
circuit in Figure~\ref{fig:dnnf} is structured according to the vtree in Figure~\ref{fig:vtree}.
For example, the and-gate highlighted in Figure~\ref{fig:dnnf} respects vtree node \(v\!=\!1\) in
Figure~\ref{fig:vtree}. Two structured DNNF circuits can be conjoined in polytime to yield a 
structured DNNF, which cannot be done on DNNF circuits
under standard complexity assumptions~\cite{darwicheJAIR02}. 

Structured decomposability, together with a stronger version of determinism, yields a class of NNF circuits
known as Sentential Decision Diagrams (SDDs)~\cite{Darwiche11}. To illustrate this stronger 
version of determinism, consider Figure~\ref{fig:sdd} and the highlighted circuit fragment. 
The fragment corresponds to the Boolean expression \((p_1 \wedge s_1) \vee (p_2 \wedge s_2) \vee (p_3 \wedge s_3)\),
where each \(p_i\) is called a {\em prime} and each \(s_i\) is called a {\em sub} (primes and subs correspond to subcircuits). 
Under any circuit input, precisely one prime will be high. In Figure~\ref{fig:sdd}, under the given circuit input,
prime \(p_2\) is high while primes \(p_1\) and \(p_3\) are low. This means that this circuit fragment, which acts
as a multiplexer, is actually passing the value of sub \(s_2\) while suppressing the value of subs \(s_1\) and \(s_3\).
As a result, the or-gate in this circuit fragment is guaranteed to be deterministic: at most one input of the or-gate
will be high under any circuit input. SDD circuits result from recursively applying this multiplexer construct,
which implies determinism, to a given vtree (the SDD circuit in
Figure~\ref{fig:sdd} is structured with respect to the vtree in Figure~\ref{fig:vtree}).
See~\cite{Darwiche11} for the formal definitions of SDD circuits and the underlying 
stronger version of determinism.\footnote{The vtree of an SDD is ordered: the distinction between 
left and right children matters.}

SDDs support polytime conjunction and disjunction.
That is, given two SDDs \(\alpha\) and \(\beta\), there is a polytime algorithm to construct another SDD \(\gamma\) 
that represents \(\alpha \wedge \beta\) or \(\alpha \vee \beta\).\footnote{If \(s\) and \(t\) are the sizes of input SDDs, 
then conjoining or disjoining the SDDs takes \(O(s \cdot t)\) time, although the resulting SDD may not be 
compressed \cite{VdBDarwiche15}. Compression is a property that ensures the canonicity of an SDD
for a given vtree~\cite{Darwiche11}.} SDDs can also be negated in linear time. The size of an SDD can 
be very sensitive to the underlying vtree, ranging from linear to exponential in some 
cases~\cite{XueChoiDarwiche12,ChoiDarwiche13}. Recall that \ems\ and \mms, the prototypical 
problems for the complexity classes $\NPPP$ and $\PPPP,$ are stated with respect to a split of variables
in the Boolean forumla. If the vtree is constrained according to this split, then these problems can be 
solved in linear time on the corresponding SDD~\cite{OztokCD16}.\footnote{A weaker condition 
exists for the class $\NPPP$ but is stated on another circuit type known as Decision-DNNF~\cite{HuangD07},
which we did not introduce in this paper; see~\cite{PipatsrisawatD09,HuangCD06}.}
Figure~\ref{fig:c-vtree} illustrates the concept of a {\em constrained vtree.}

\begin{figure}[tb]
  \centering
  \includegraphics[width=.8\linewidth]{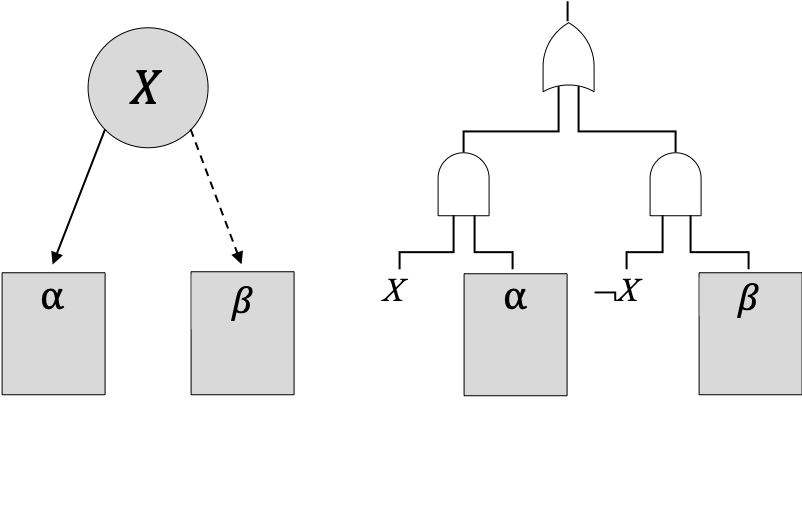}
  \caption{OBDD fragment and corresponding NNF circuit. \label{fig:obdd2nnf}}
    \Description{Ordered Binary Decision Diagrams (OBDDs).}
\end{figure}

SDDs subsume the well known Ordered Binary Decision Diagrams (OBDDs) and are 
exponentially more succinct than OBDDs~\cite{Bova16}; see also~\cite{DBLP:journals/mst/BolligB19}. 
An OBDD is an ordered decision graph:
the variables on every path from the root to a leaf (\(0\) or \(1\)) respect a given variable
order~\cite{Bryant86,MeinelT98,Wegener00}; see Figure~\ref{fig:obdd}. An OBDD node and its corresponding NNF 
fragment are depicted in Figure~\ref{fig:obdd2nnf}. As the figure shows, this fragment is also 
a multiplexer as in SDDs, except that we have precisely two primes which correspond to a 
variable and its negation. When an SDD is structured with respect to a {\em right-linear vtree,} 
the result is an OBDD; see Figure~\ref{fig:rl-vtree}. SDDs and OBDDs
can also be contrasted based on how they make decisions. An OBDD makes a decision
based on the state of a {\em binary} variable and hence its decisions are always binary.
An SDD makes decisions based on {\em sentences} (primes) so the corresponding
number of decisions (subs) is not restricted to being binary. 

The compilation of Boolean formula into tractable NNF circuits is done by systems known
as {\em knowledge compilers.} Examples include \ctd\footnote{\url{http://reasoning.cs.ucla.edu/c2d}}~\cite{Darwiche04},
\cudd\footnote{https://davidkebo.com/cudd},
\mctd\footnote{\url{http://reasoning.cs.ucla.edu/minic2d}}~\cite{OztokD14b,OztokD18}, 
\df\footnote{\url{http://www.cril.univ-artois.fr/kc/d4}}~\cite{DBLP:conf/ijcai/LagniezM17},
\dsharp\footnote{\url{https://bitbucket.org/haz/dsharp}}~\cite{DBLP:conf/ai/MuiseMBH12} and
the \sdd\ library~\cite{ChoiDarwiche13}.\footnote{\url{http://reasoning.cs.ucla.edu/sdd} (open source). 
A Python wrapper of the \sdd\ library is available at \url{https://github.com/wannesm/PySDD}
and an NNF-to-SDD circuit compiler, based on PySDD, is available open-source at \url{https://github.com/art-ai/nnf2sdd}}
See also \url{http://beyondnp.org/pages/solvers/knowledge-compilers/}.
A connection was made between model counters and knowledge compilers in~\cite{HuangD07}, showing how 
model counters can be turned into knowledge compilers by keeping a {\em trace} of the exhaustive search
they conduct on a Boolean formula. The \dsharp\ compiler for d-DNNF~\cite{DBLP:conf/ai/MuiseMBH12} was 
the result of keeping a trace of the \sharps\ model counter~\cite{DBLP:conf/sat/Thurley06}. While
dedicated SAT solvers remain the state of the art for solving \sat, the state of the art for (weighted)
model counting are either knowledge compilers or model counters whose traces are
(subsets) of d-DNNF circuits.

\begin{figure}[tb]
  \centering
  \includegraphics[width=.8\linewidth]{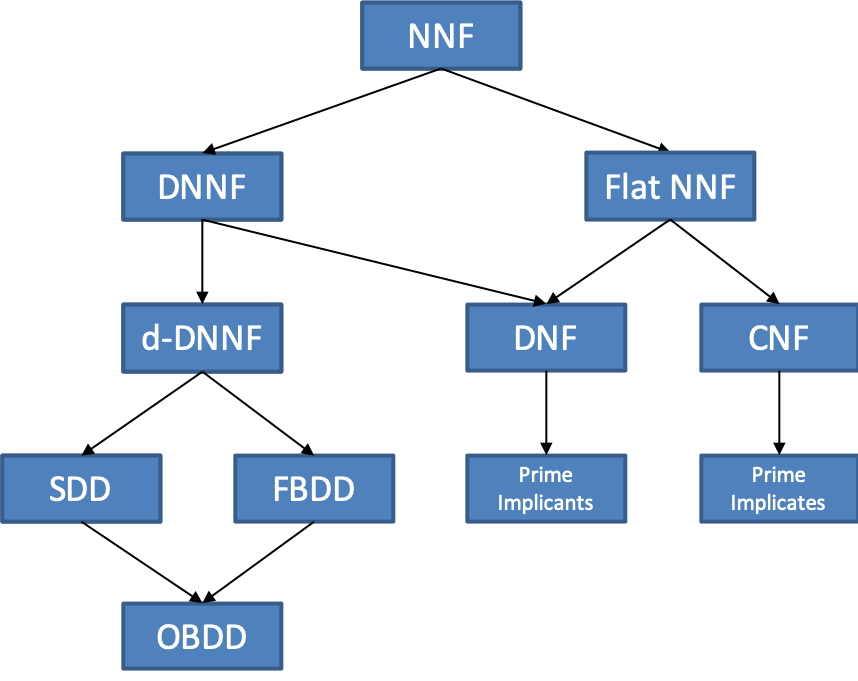}
  \caption{A partial taxonomy of NNF circuits. \label{fig:languages}}
    \Description{Taxonomy of NNF circuits.}
\end{figure}

Figure~\ref{fig:languages} depicts a partial taxonomy of NNF circuits, most of which are tractable.
A more comprehensive taxonomy was studied in~\cite{darwicheJAIR02} under the term
{\em knowledge compilation map.} A recent update on this map can
be found in~\cite{kocoon19FS}, which includes recent results on the exponential separations
between tractable NNF circuits. 

Tractable NNF circuits were connected to, utilized by, and also benefited from other areas of
computer science. This includes database theory,
where connections were made to the notions of provenance and lineage, e.g.,~\cite{kocoon19AA,DBLP:journals/mst/JhaS13}. 
It also includes communication complexity, which led to sharper separation
results between tractable NNF circuits, e.g.,~\cite{DBLP:conf/ijcai/BovaCMS16,DBLP:conf/uai/BeameL15}. 
Further applications to probabilistic reasoning and machine learning include the utilization of
 tractable circuits for uniform sampling~\cite{DBLP:conf/lpar/SharmaGRM18} and improving deep learning 
 through the injection of symbolic knowledge; see, e.g.,~\cite{DBLP:conf/icml/XuZFLB18,DBLP:conf/nips/XieXMKS19}.

\section{Logic For Learning From Data and Knowledge}
\label{sec:bk}

\begin{figure}[tb]
  \centering
  \includegraphics[width=\linewidth]{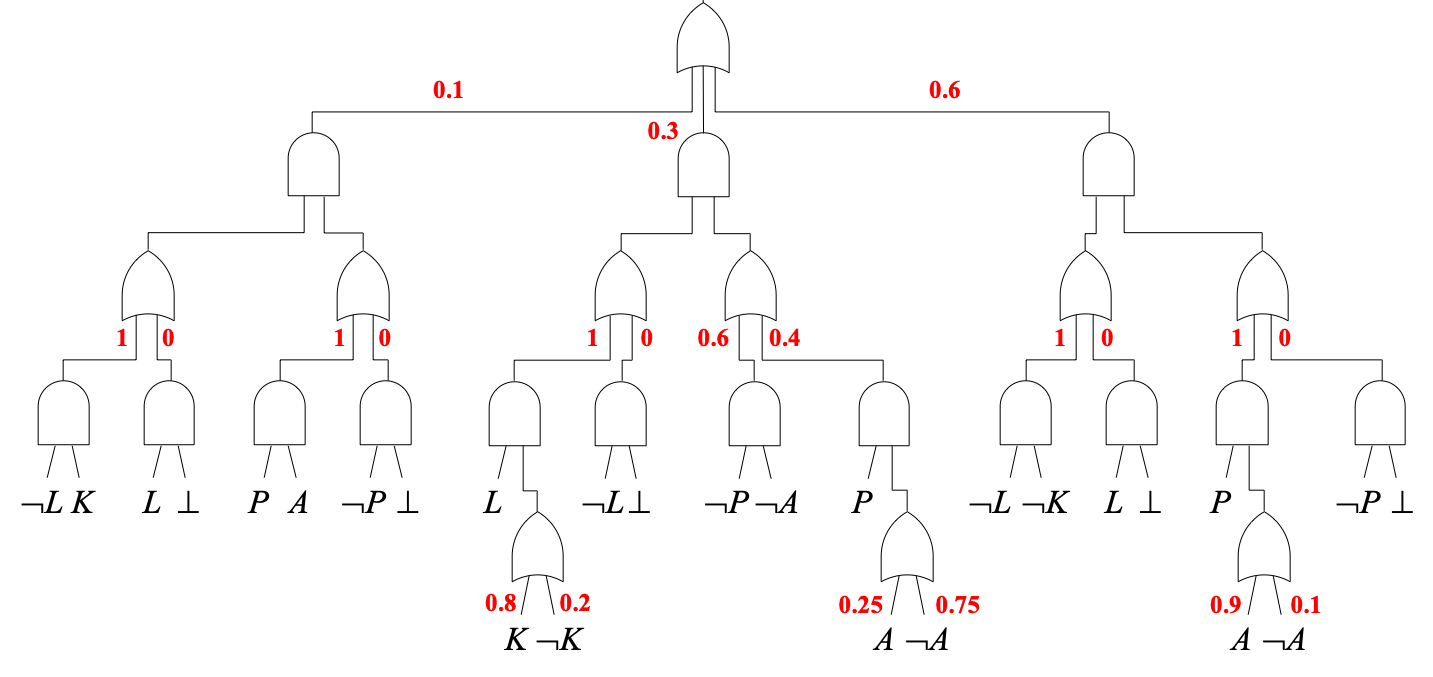}
  \caption{A probabilistic SDD circuit. The probabilities annotating inputs
  of or-gate are known as PSDD {\em parameters.} \label{fig:psdd}}
    \Description{Probabilistic Sentential Decision Diagrams (PSDDs).}
\end{figure}

The second role we consider for logic is in learning distributions from a combination of data and 
symbolic knowledge. 
This role has two dimensions: representational and computational. On the representational side,
we use logic to eliminate situations that are impossible according to symbolic knowledge,
which can reduce the amount of needed data and increase the robustness of learned representations.
On the computational side, we use logic to factor the space of possible situations into a tractable NNF circuit
so we can learn a distribution over the space and reason with it efficiently. 

Consider the SDD circuit in Figure~\ref{fig:psdd} which we showed earlier to have \(9\)
satisfying inputs out of \(16\). Suppose that our goal is to induce a distribution on this space of satisfying
circuit inputs. We can do this by simply assigning a distribution to each or-gate in the circuit as shown
in Figure~\ref{fig:psdd}: each input of an or-gate is assigned a probability while ensuring that the
probabilities assigned to the inputs add up to \(1\). These local distributions are independent of one 
another. Yet, they are guaranteed to induce a normalized distribution over the space of satisfying 
circuit inputs. The resulting circuit is known as a {\em probabilistic SDD} circuit or a PSDD for short~\cite{KisaKR14}.

\begin{figure}[tb]
  \centering
  \includegraphics[width=\linewidth]{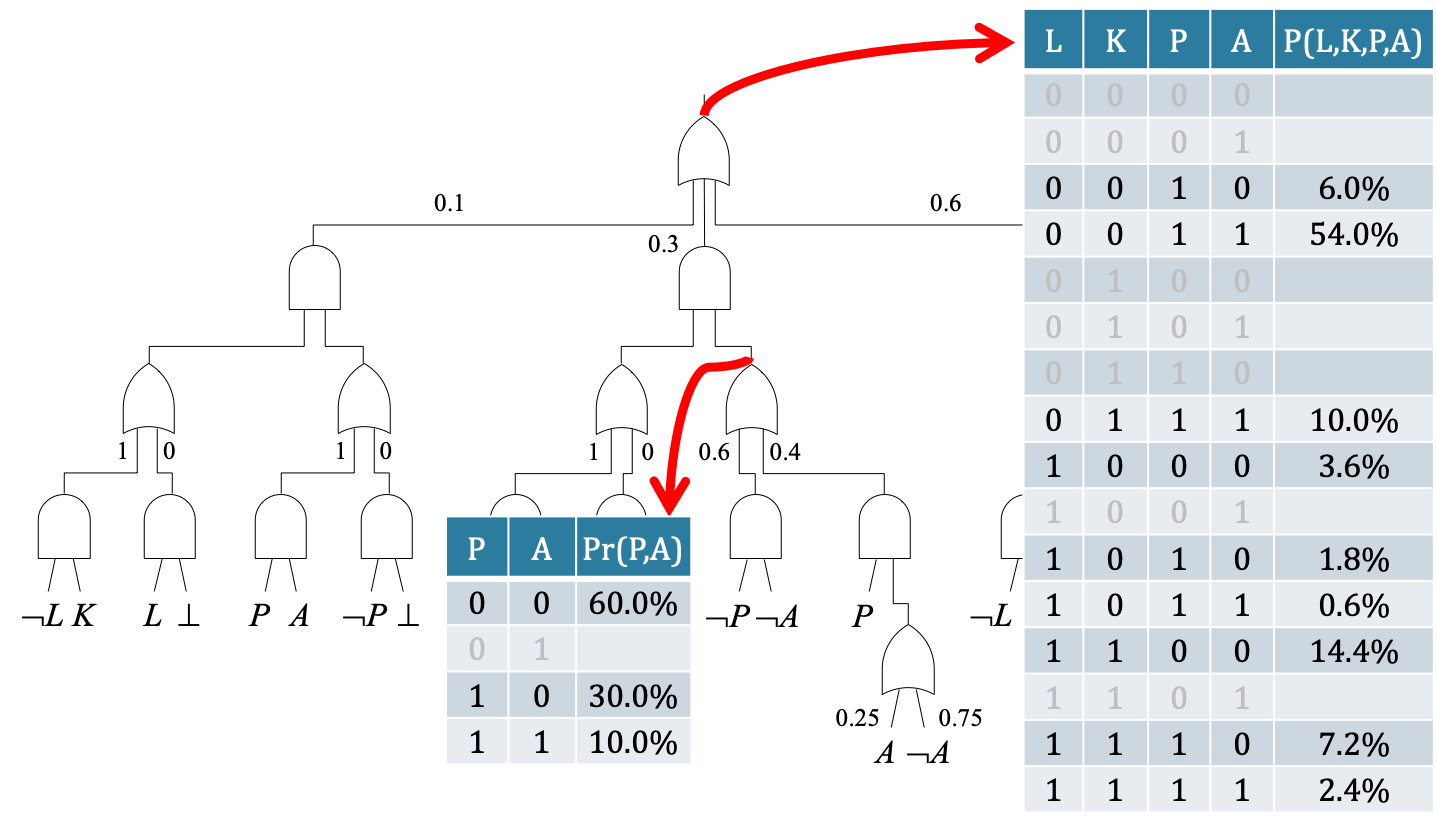}
  \caption{The compositional distributions of a PSDD circuit. \label{fig:dist}}
    \Description{Semantics of PSDDs.}
\end{figure}

Given a PSDD with variables \(\X\), the distribution \(\pr(\X)\) it induces can be 
obtained as follows. To compute the probability of input \(\x\) we perform a bottom-up pass
on the circuit while assigning a value to each literal and gate output. 
The value of a literal is just its value in input \(\x\) (\(1\) or \(0\)). 
The value of an and-gate is the product of values assigned to its inputs.  
The value of an or-gate is the weighted sum of values assigned to its inputs (the 
weights are the probabilities annotating the gate inputs).
The value assigned to the circuit output will then be the probability \(\pr(\x)\).
Figure~\ref{fig:dist} depicts the result of applying this evaluation process. As the figure shows,
the probabilities of satisfying circuit inputs add up to \(1\). Moreover, the probability of each unsatisfying
input is \(0\). The PSDD has a compositional semantics: each or-gate induces a distribution
over the variables in the subcircuit rooted at the gate. Figure~\ref{fig:dist} depicts the distribution
induced by an or-gate over variables \(P\) and \(A\).

\begin{figure}[tb]
  \centering
  \includegraphics[width=\linewidth]{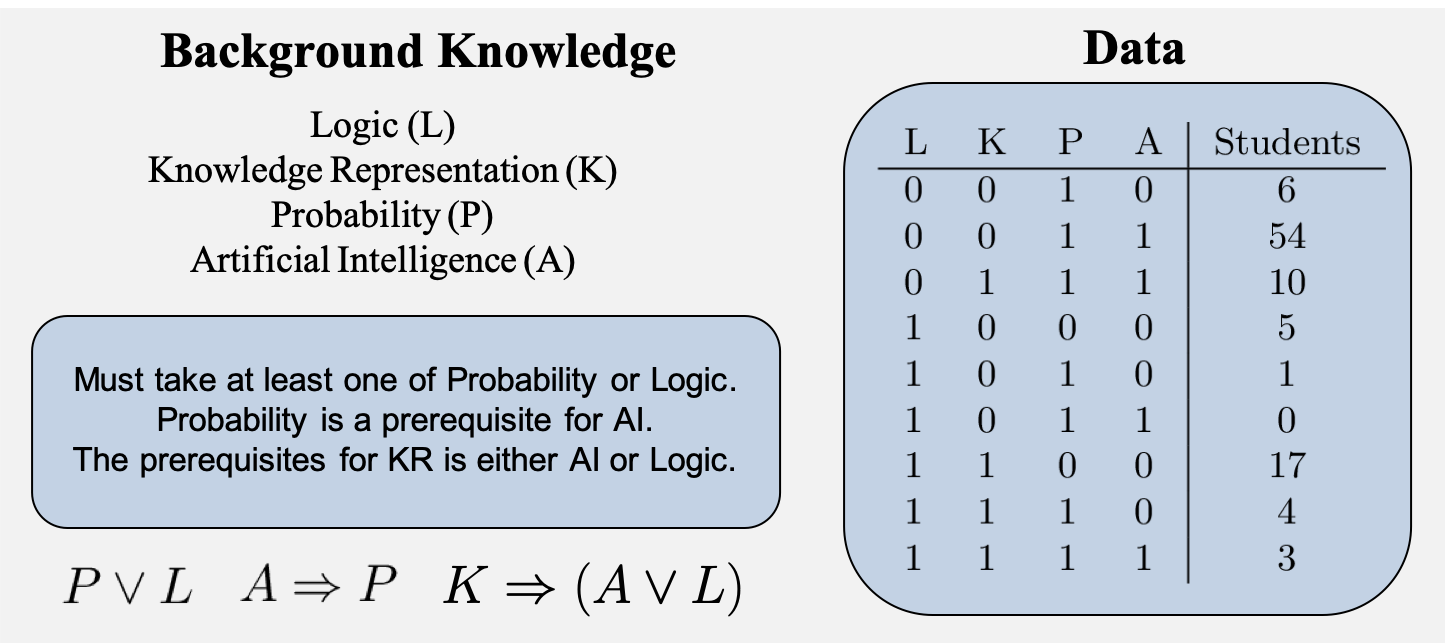}
  \caption{Learning from data and knowledge. \label{fig:bk}}
    \Description{Learning from knowledge and data.}
\end{figure}

To see how PSDDs can be used to learn from a combination of data and symbolic knowledge,
consider the example in Figure~\ref{fig:bk} from~\cite{KisaKR14}. What we have here is a dataset 
that captures student enrollments in four courses offered by a computer science department. 
Each row in the dataset represents the number of students who have enrolled in the corresponding 
courses. Our goal is to learn a distribution from this data and the knowledge we have about the 
course prerequisites and requirements that is shown in Figure~\ref{fig:bk}. 

This symbolic knowledge can be captured by the propositional statement 
\((P \vee L) \wedge (A \Rightarrow P) \wedge (K \Rightarrow (A \vee L))\). The first
step is to compile this knowledge into an SDD circuit. The SDD circuit in Figure~\ref{fig:psdd}
is actually the result of this compilation (using the vtree in Figure~\ref{fig:vtree}).
This SDD circuit will evaluate to \(1\) for any course combination that is allowed by the 
prerequisites and requirements, otherwise it will evaluate to \(0\). Our next step is to induce 
a distribution on the satisfying inputs of this course, that is, the valid course combinations. 
This can be done by learning a distribution for each or-gate in the SDD circuit, to yield a PSDD circuit. 

The data in Figure~\ref{fig:bk} is {\em complete} in the sense that each row specifies precisely
whether a course was enrolled into or not by the corresponding number of students. An {\em incomplete}
dataset in this case would, for example, specify that \(30\) students took logic, AI and probability,
without specifying the status of enrollment in knowledge representation. If the data is complete,
the {\em maximum-likelihood} parameters of a PSDD can be learned in time linear in the PSDD
size. All we need to do is evaluate the SDD circuit for each example in the dataset, while keeping
track of how many times a wire become high; see~\cite{KisaKR14} for details. The parameters
shown in Figure~\ref{fig:psdd} were actually learned using this procedure so the distribution
they induce is the one guaranteed to maximize the probability of given data (under the 
chosen vtree).\footnote{See \url{https://github.com/art-ai/pypsdd} for a package that
learns PSDDs and \url{http://reasoning.cs.ucla.edu/psdd/} for additional tools.}

Both \mpe\ and \mar\ queries, discussed in Section~\ref{sec:computation}, can be computed
in time linear in the PSDD size~\cite{KisaKR14}. Hence, not only can we estimate parameters
efficiently under complete data, but we can also reason efficiently with the learned distribution. 
Finally, the PSDD is a complete and canonical representation of probability distributions. That is, 
PSDDs can represent any distribution, and there is a unique PSDD for that distribution (under 
some conditions)~\cite{KisaKR14}.

PSDD circuits are based on the stronger versions of decomposability and determinism that underly
SDD circuits. Probabilistic circuits that are based on the standard properties of decomposability 
and determinism are called ACs (Arithmetic Circuits)~\cite{Darwiche03}. Those based on
decomposability only were introduced about ten years later and are known as SPNs (Sum-Product Networks)~\cite{PoonD11}. 
A treatment on the relative tractability and succinctness of these three circuit types can be 
found in~\cite{ChoiDarwiche17,ShenCD16}.

\subsection{Learning With Combinatorial Spaces}

\begin{figure}[tb]
  \centering
  \includegraphics[width=.9\linewidth]{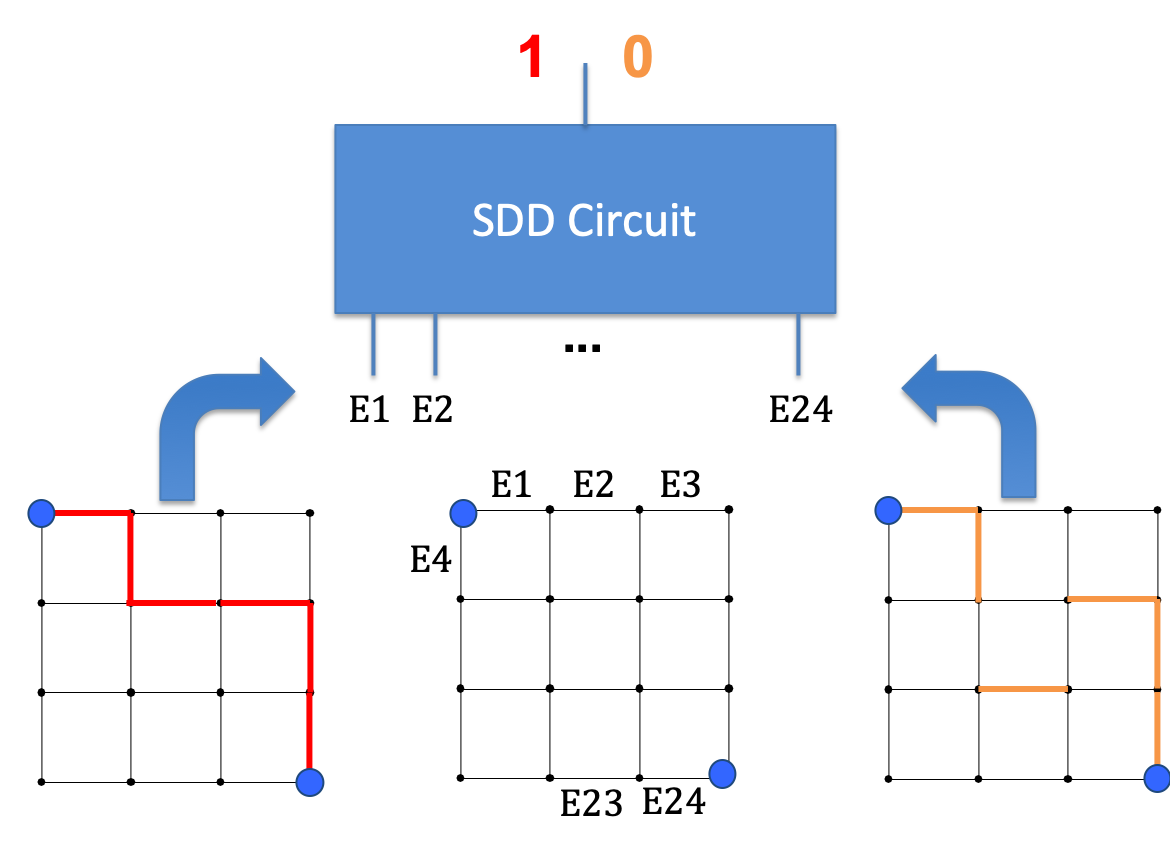}%
  \caption{Encoding routes using SDDs. The red variable assignment (left) encoudes
  a valid route. The orange variable assignment (right) does not encode a valid 
  route as the edges are disconnected. \label{fig:comb-routes}}
      \includegraphics[width=\linewidth]{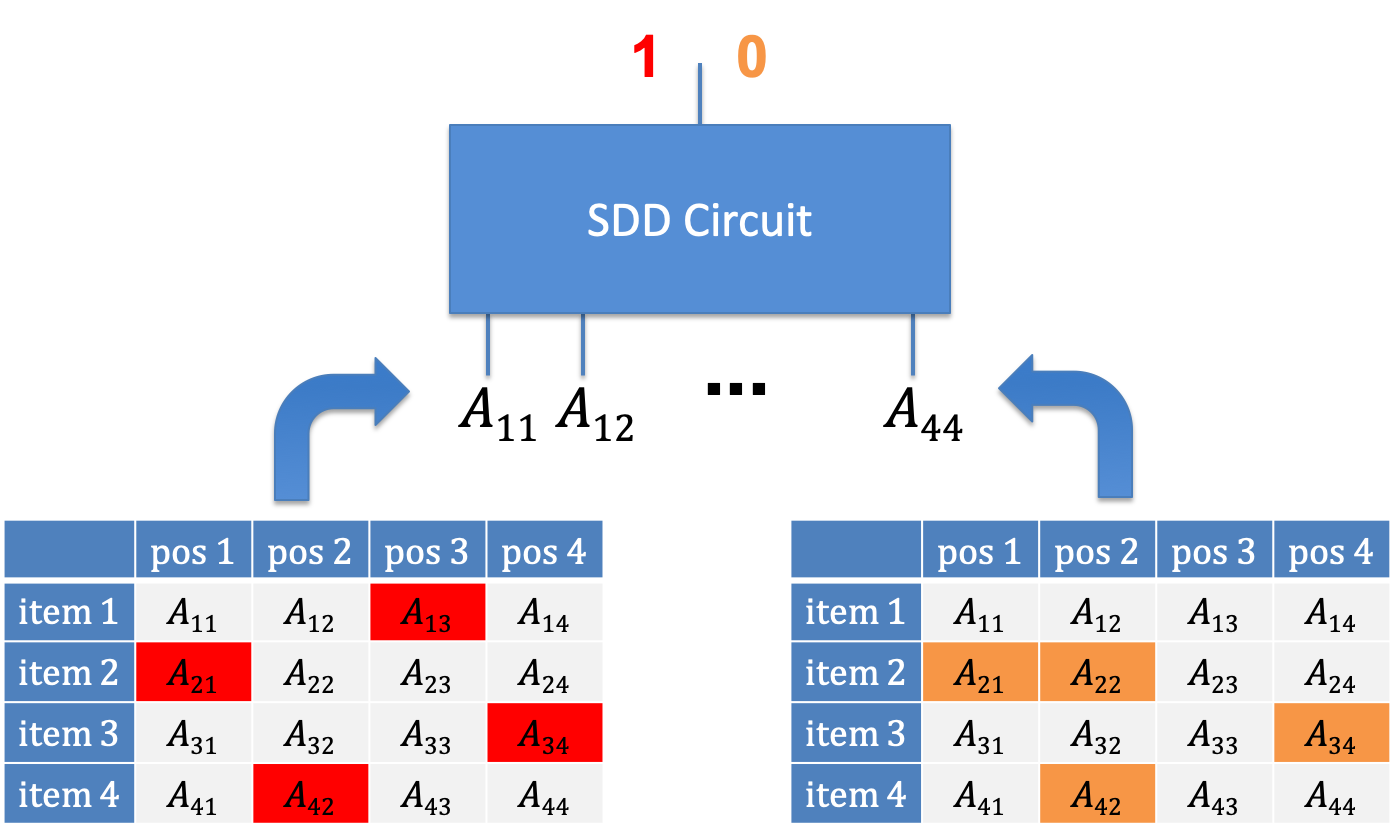}
  \caption{Encoding rankings (total orderings) using SDDs. 
  The red variable assignment (left) encodes a valid ranking. The orange 
  variable assignment (right) does not encode a valid ranking (e.g.,
  item~\(2\) appears in two positions). \label{fig:comb-rankings}}
    \Description{Encoding combinatorial objects using SDDs.}
\end{figure}

We now discuss another type of symbolic knowledge that arises when learning distributions over combinatorial objects.

Consider Figures~\ref{fig:comb-routes} and~\ref{fig:comb-rankings} which depicts two common 
types of combinatorial objects: routes and total orderings. In the case of routes, we have a map
(in this case a grid for simplicity), a source and a destination. Our goal is to learn a
distribution over the possible routes from the source to the destination given a dataset of routes,
as discussed in~\cite{ChoiTavabiDarwiche16}. In the case of total orderings (rankings), we have \(n\) 
items and we wish to learn a distribution over the possible rankings of these 
items, again from a dataset of rankings, as discussed in~\cite{ChoiVdbDarwiche15}.

It is not uncommon to develop dedicated learning and inference algorithms when dealing
with combinatorial objects. For example, a number of dedicated frameworks exist for
learning and inference with distributions over rankings; see, e.g.,~\cite{Mallows,fligner1986distance,meila2010dirichlet}. 
What we will do, however, is show how learning and inference with combinatorial objects including rankings
can be handled systematically using tractable circuits as proposed
in~\cite{ChoiVdbDarwiche15,ChoiTavabiDarwiche16,ShenChoiDarwiche17,ChoiSD17,DBLP:conf/aaai/ShenGDC19}.

Consider the grid in Figure~\ref{fig:comb-routes} (more generally, a map is modeled using an undirected graph). 
We can represent each edge \(i\) in the map by a Boolean variable \(E_i\) and each route as a variable 
assignment that sets only its corresponding edge variables to true. While every route can be represented
by a variable assignment (e.g., the red route on the left of Figure~\ref{fig:comb-routes}), some assignments 
will correspond to invalid routes (e.g., the orange one on the right of Figure~\ref{fig:comb-routes}).
One can construct a Boolean formula over edge variables whose satisfying assignments correspond
precisely to valid (connected) routes and include additional constraints to keep only, for example,
simple routes with no cycles; see \cite{ChoiTavabiDarwiche16,NishinoYMN17a} for how this can be done.

After capturing the space of valid routes using an appropriate Boolean formula, we compile it to an SDD circuit.
Circuit inputs that satisfy the SDD will then correspond to the space of valid routes; see Figure~\ref{fig:comb-routes}.
A complete dataset in this case will be a multi-set of variable assignments, each corresponding to a taken route
(obtained, for example, from GPS data). We can then learn PSDD parameters from this dataset and 
compiled SDD as done in~\cite{ChoiTavabiDarwiche16,ChoiSD17}.

Figure~\ref{fig:comb-rankings} contains another example of applying this approach to learning distributions
over rankings with \(n\) items. In this case, we use \(n^2\) Boolean variables \(A_{ij}\) to encode a ranking,
by setting variable \(A_{ij}\) to true iff item~\(i\) is in position~\(j\). One can also construct a
Boolean formula whose satisfying assignments correspond precisely to valid rankings. Compiling the
formula to an SDD and then learning PSDD parameters from data allow us to learn a distribution over
rankings that can be reasoned about efficiently. This proposal for rankings (and partial rankings, including
tiers) was extended in~\cite{ChoiVdbDarwiche15}, showing competitive results with some dedicated
approaches for learning distributions over rankings. It also included an account for learning PSDD
parameters from incomplete data and from {\em structured data} in which examples are expressed using
arbitrary Boolean formula instead of just variable assignments. 

A combinatorial probability space is a special case of the more general {\em structured probability
space,} which is specified by a Boolean formula (i.e., the satisfying assignments need not correspond to
combinatorial objects). The contrast is a standard probability space that is defined over all instantiations 
of a set of variables. We will use the term {\em structured} instead of {\em combinatorial} in the 
next section, as adopted in earlier 
works~\cite{ChoiVdbDarwiche15,ChoiTavabiDarwiche16,ChoiSD17,DBLP:conf/aaai/ShenGDC19},
even as we continue to give examples from combinatorial spaces.

\subsection{Conditional Spaces}

We now turn to the notion of a {\em conditional space:} a structured space that is determined by the state 
of another space. The interest is in learning and reasoning with distributions over conditional spaces.
This is a fundamental notion that arises in the context of causal probabilistic models, which require such conditional distributions
when specifying probabilistic relationships between causes and effects. More generally, it arises in directed
probabilistic graphical models, in which the graph specifies conditional independence relationships 
even though it may not have a causal interpretation.

\begin{figure}[tb]
  \centering
  \includegraphics[width=.50\linewidth]{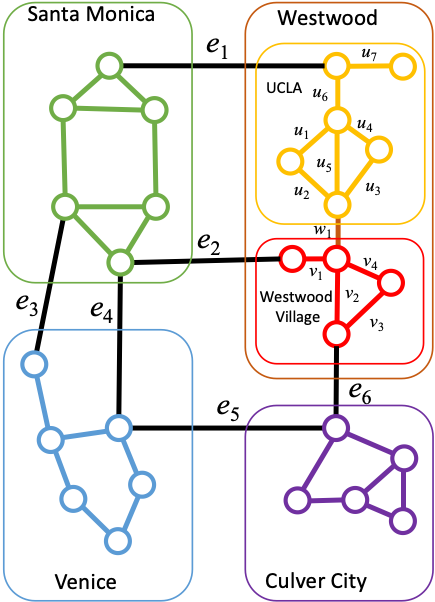}
   \caption{A three-level hierarchical map. \label{fig:hmap}}
     \Description{Hierarchical map.}
 \end{figure}

A concrete example comes from the notion of a {\em hierarchical map}, which was introduced to better 
scale the compilation of maps into tractable circuits~\cite{ChoiSD17,DBLP:conf/aaai/ShenGDC19}. 
Figure~\ref{fig:hmap} depicts an example of a three-level hierarchical map with a number of regions. 
It is a simplified map of neighborhoods in the Los Angeles Westside, where edges 
represent streets and nodes represent intersections. Nodes of the LA Westside have been 
partitioned into four sub-regions: Santa Monica, Westwood, Venice and Culver City. 
Westwood is further partitioned into two sub-regions: UCLA and Westwood Village. 

\begin{figure}[tb]  
   \includegraphics[width=.60\linewidth]{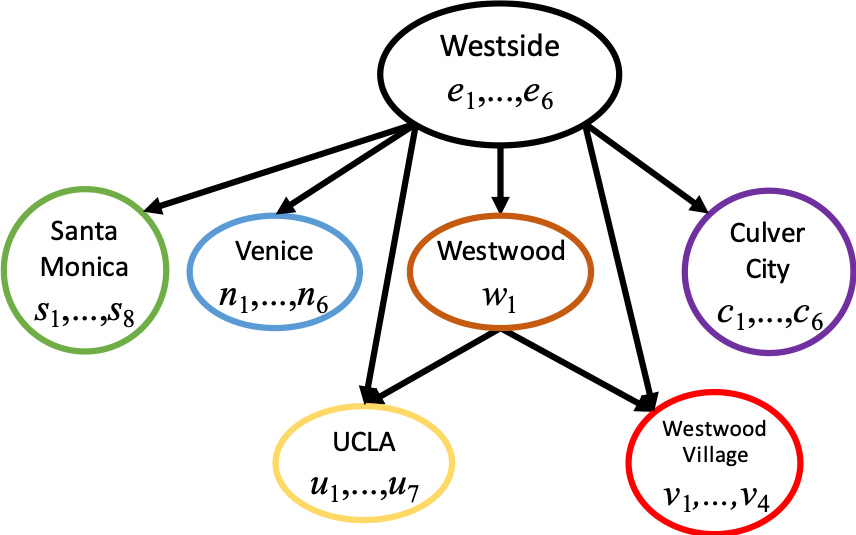}
  \caption{Specifying conditional independence relationships using a directed acyclic graph.
The relationship between a node and its parent requires a conditional space. \label{fig:cdag}}
  \Description{Structured Bayesian network (SBN).}
\end{figure}

The main intuition behind a hierarchical map is that navigation behavior in a region \(R\) 
can become independent of navigation behavior in other regions, once we know how region \(R\) was
entered and exited. These independence relations are specified using a directed acyclic graph as shown
in Figure~\ref{fig:cdag} (called a {\em cluster DAG} in~\cite{DBLP:conf/aaai/ShenCD18}).
Consider the root node `Westside' which contains variables \(e_1, \ldots, e_6\). These variables
represent the roads used to cross between the four (immediate) sub-regions of the Westside.
Once we know the state of these variables, we also know how each of these regions may have been entered
and exited so their inner navigation behaviors become independent of one another.

\begin{figure}[tb]
  \centering
  \includegraphics[width=.85\linewidth]{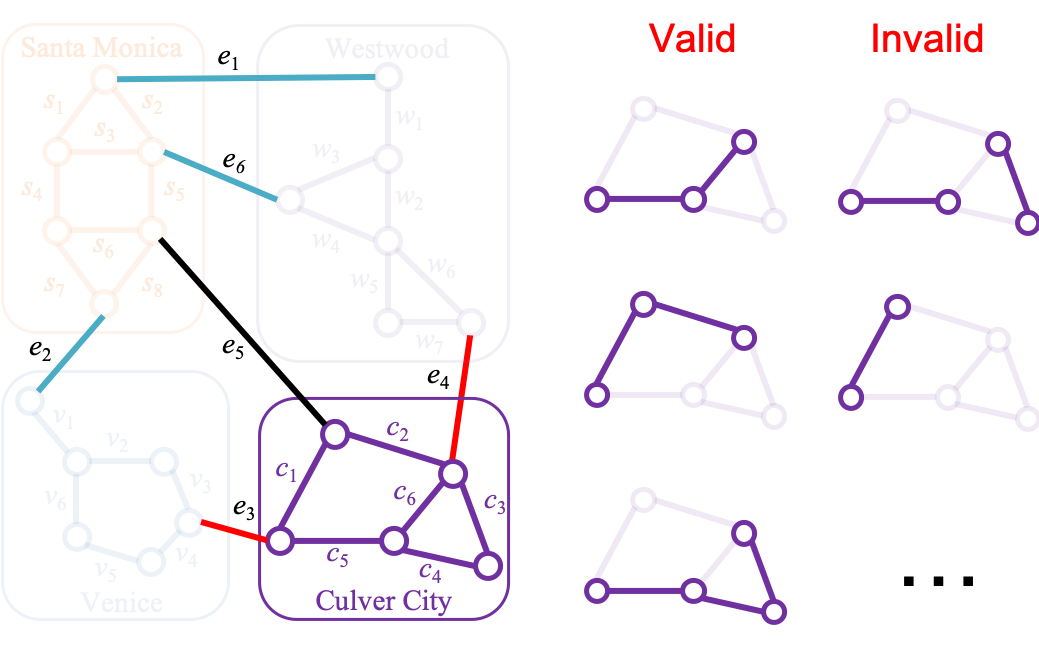}
  \caption{Illustrating the notion of a conditional space. The valid
  routes inside Culver City are a function of the Westside crossings we
  use to enter and exit the city. \label{fig:cond}}
    \Description{Conditional spaces and structured probability spaces.}
\end{figure}

Let us now consider Figure~\ref{fig:cond} to see how the notion of a conditional space arises in this context.
The figure highlights Culver City with streets \(e_1, \ldots, e_6\): the ones for crossing between regions
in the Westside. What we need is a structured space over inner roads \(c_1, \ldots, c_6\) of Culver City
that specifies valid routes inside the city. But this structured space depends on how we enter and exit the city, 
which is specified by another space over crossings \(e_1, \ldots, e_6\). 
That is, the structured space over Culver City roads is conditional on the space over Westside crossings.

The left of Figure~\ref{fig:cond} expands this example by illustrating the structured space
over \(c_1, \ldots, c_6\) assuming we entered/existed Culver city using crossings \(e_3\) and \(e_4\) (highlighted
in red). The illustration shows some variable assignments that belong to this structured space (valid)
and some that do not (invalid). If we were to enter/exit Culver city using, say, crossings \(e_3\) and \(e_5\), then
the structured space over \(c_1, \ldots, c_6\) would be different.

\begin{figure}[tb]
  \centering
  \includegraphics[width=.85\linewidth]{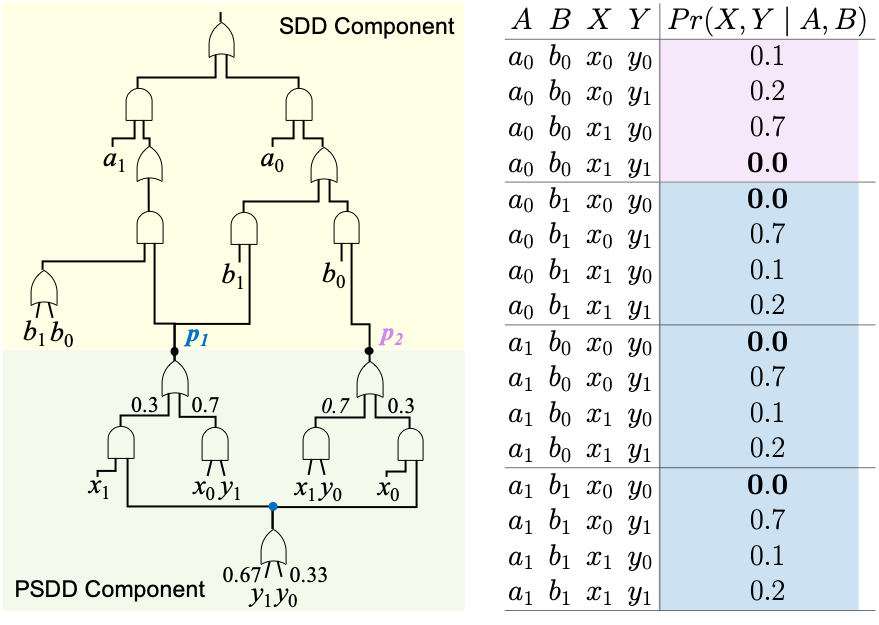}
  \caption{A conditional PSDD and the two conditional distributions it represents. \label{fig:c-psdd}}
    \Description{Conditional PSDD.}
\end{figure}

Figure~\ref{fig:c-psdd} depicts a new class of tractable circuits, called {\em conditional PSDDs,} 
which can be used to induce distributions on conditional spaces~\cite{DBLP:conf/aaai/ShenCD18}.
In this example, we have a structured space over variables \(X,Y\) that is conditioned on a 
space over variables \(A, B\). The conditional PSDD has two components: 
an SDD circuit (highlighted in yellow) and a multi-rooted PSDD (highlighted in green). The conditional
distributions specified by this conditional PSDD are shown on the right of Figure~\ref{fig:c-psdd}.
There are two of them: one for state \(a_0,b_0\) of variables \(A,B\) and another for the remaining
states. The structured space of the first distribution corresponds to the Boolean formula 
\(x_0 \vee y_0\). The structured space for the second distribution corresponds to \(x_1 \vee y_1\).

\begin{figure*}[h]
  \centering
  \includegraphics[width=.8\linewidth]{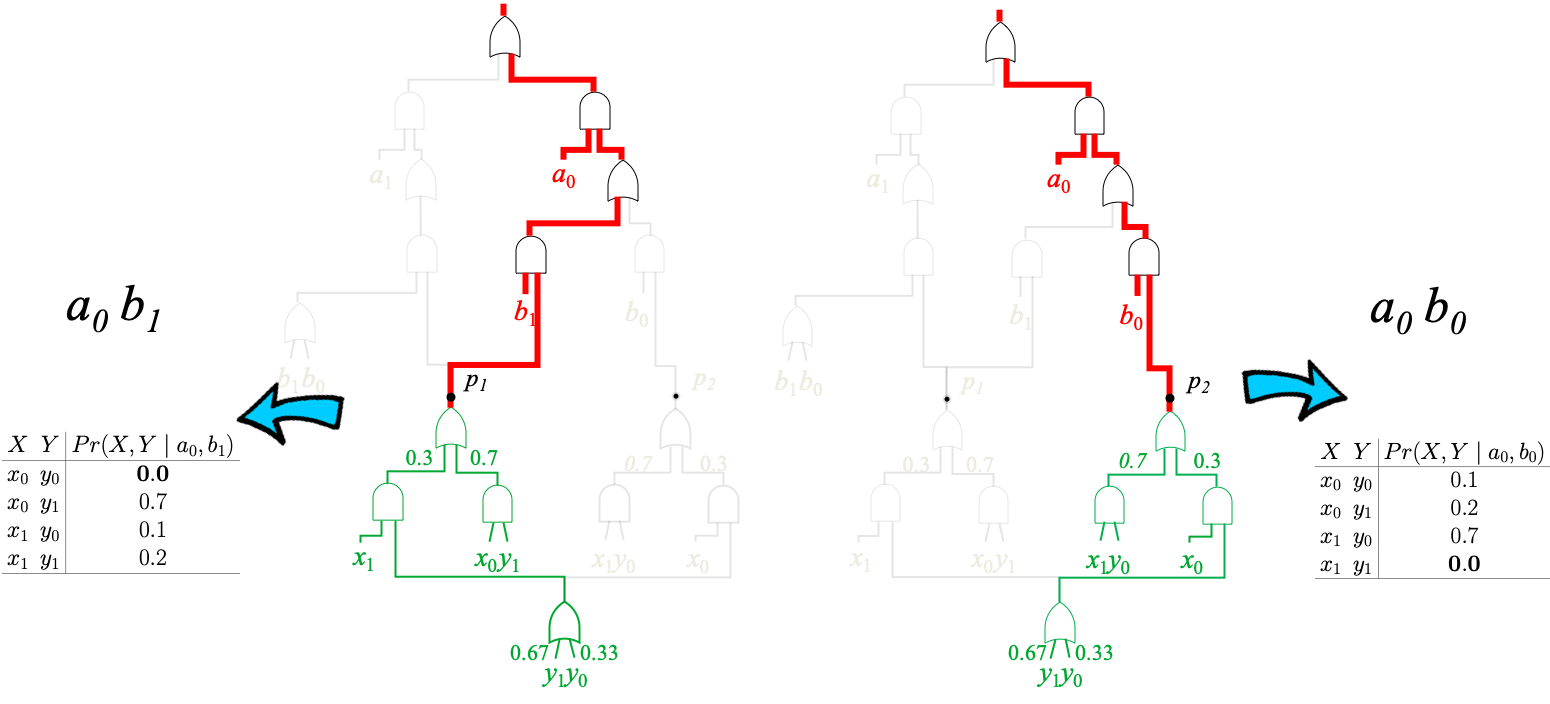}
  \caption{Selecting conditional distributions using the conditional PSDD of Figure~\ref{fig:c-psdd}. \label{fig:scpt}}
    \Description{Semantics of conditional PSDDs.}
\end{figure*}

The semantics of a conditional PSDD is relatively simple and illustrated in Figure~\ref{fig:scpt}.
Consider state \(a_0,b_0\) of variables \(A,B\) (right of Figure~\ref{fig:scpt}). Evaluating the SDD 
component at this input leads to selecting the PSDD rooted at \(p_2\), which generates the 
distribution conditioned on this state. Evaluating the SDD at any other state of variables \(A,B\) leads
to selecting the PSDD rooted at \(p_1\), which generates the distribution for these 
states (left of  Figure~\ref{fig:scpt}).

\begin{figure}[h]
  \centering
  \includegraphics[width=.75\linewidth]{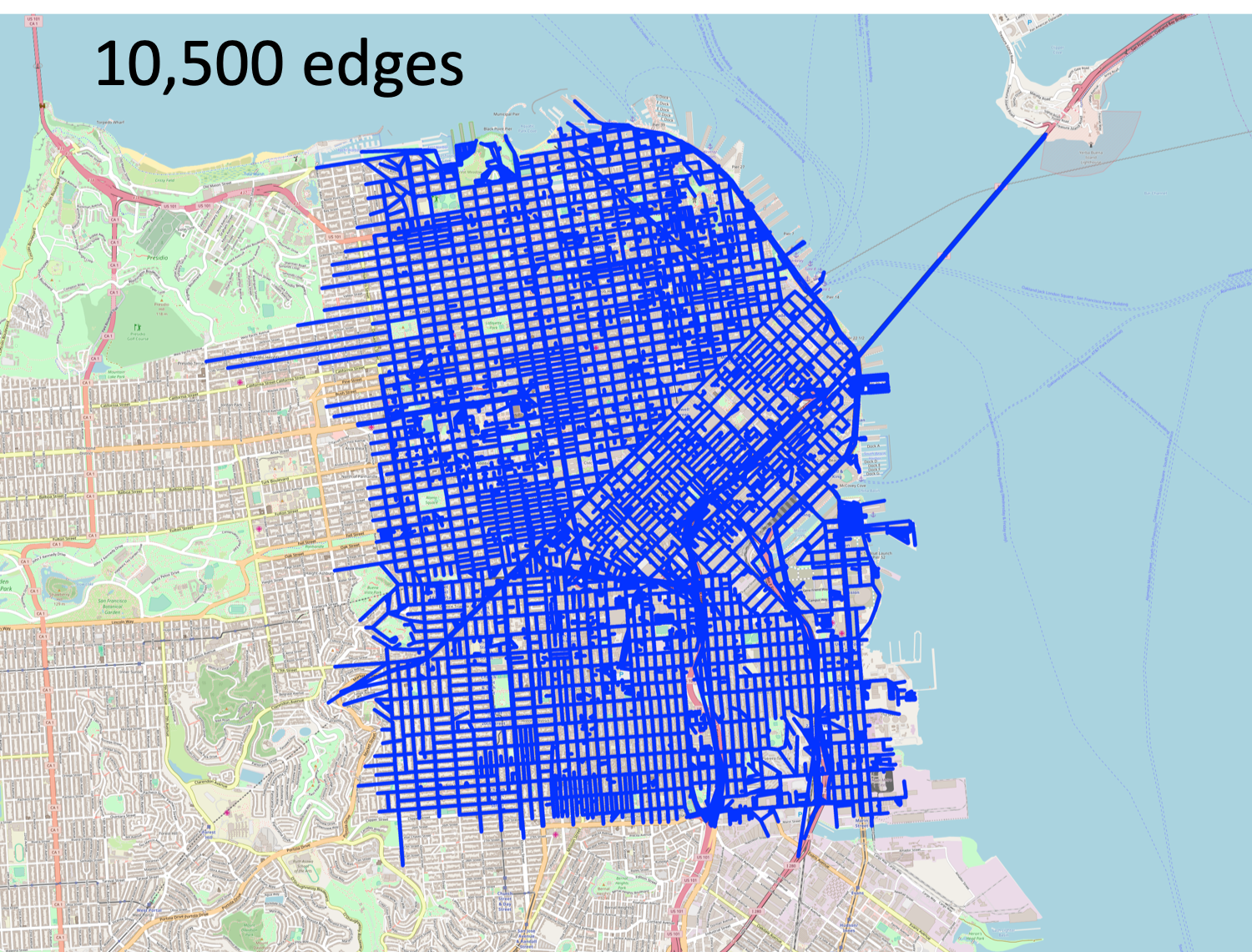}
  \caption{A map of downtown San Fransisco that was compiled into SDD/PSDD circuits. \label{fig:sf}}
    \Description{Compiling maps into tractable circuits.}
\end{figure}

\begin{figure}[tb]
  \centering
  \includegraphics[width=\linewidth]{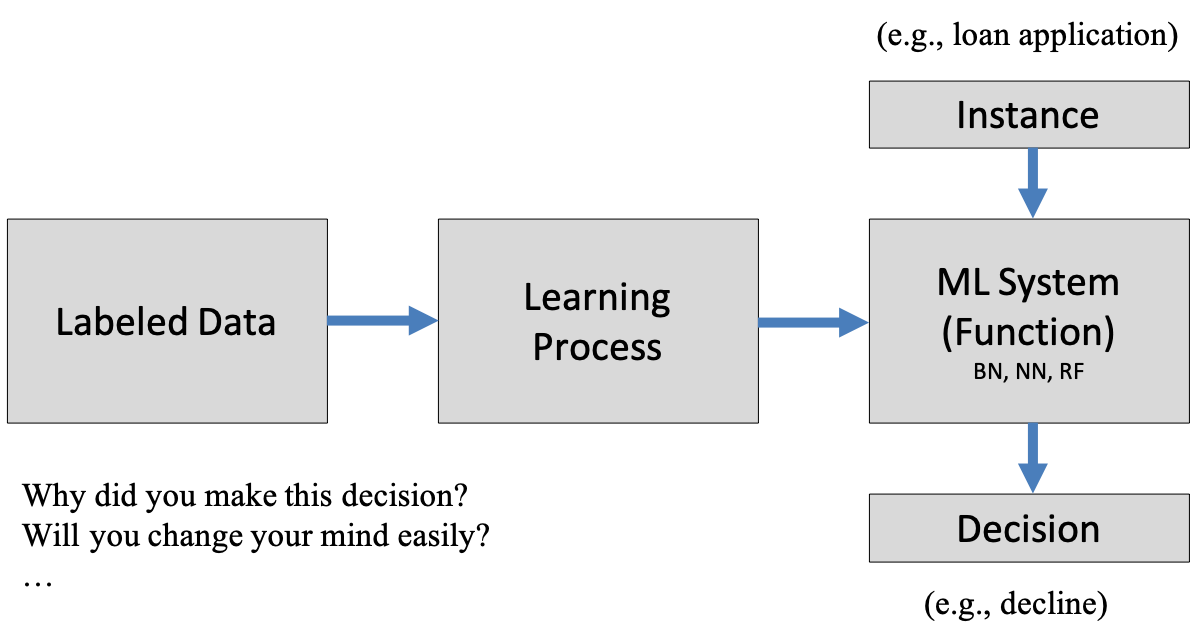}
  \caption{Reasoning about machine learning systems. \label{fig:ml}}
    \Description{Explainable AI.}
\end{figure}

When a cluster DAG such as the one in Figure~\ref{fig:cdag} is quantified using conditional
PSDDs, the result is known as a {\em structured Bayesian network (SBN)}~\cite{DBLP:conf/aaai/ShenCD18}. 
The conditional PSDDs of an SBN can be multiplied together to yield a classical PSDD
over network variables~\cite{ShenCD16}. To give a sense of practical complexity, the map of San Francisco 
depicted in Figure~\ref{fig:sf} has \(10,500\) edges. A corresponding hierarchical map  
used in~\cite{DBLP:conf/aaai/ShenGDC19} was compiled into a PSDD with size of 
about \(8.9\)M (edges). The parameters of this SDD were learned from routes collected from GPS
data and the induced distribution was used for several reasoning and classification 
tasks~\cite{DBLP:conf/aaai/ShenGDC19}.\footnote{See \url{https://github.com/hahaXD/hierarchical_map_compiler} 
for a package that compiles route constraints into an SDD.}

\section{Logic For Meta Reasoning}
\label{sec:meta}

We now turn to a most recent role for logic in AI: Reasoning about the behavior of machine
learning systems.

Consider Figure~\ref{fig:ml} which depicts how most machine learning systems are constructed today. 
We have a labeled dataset that is used to learn a classifier, which is commonly a neural
network, a Bayesian network classifier or a random forest. These classifiers are effectively functions
that map instances to decisions. For example, an instance could be a loan application and the decision
is whether to approve or decline the loan. There is now considerable interest in reasoning about the
behavior of such systems. Explaining decisions is at the
forefront of current interests: Why did you decline Maya's application? Quantifying the robustness of these 
decisions is also attracting a lot of attention: Would reversing the decision on Maya require
many changes to her application? 
In some domains, one expects the learned systems to satisfy certain properties, like monotonicity, 
and there is again an interest in proving such properties formally. For example, can we 
guarantee that a loan applicant will be approved when the only difference they have with
another approved applicant is their higher income?
These interests, however, are challenged by the numeric nature of machine learning systems and the
fact that these systems are often model-free, e.g., neural networks, so they appear as black
boxes that are hard to analyze. 

The third role for logic we discuss next rests on the following observation: Even though these machine learning classifiers
are learned from data and numeric in nature, they often implement discrete decision functions.
One can therefore extract these decisions functions and represent them symbolically using tractable circuits.
The outcome of this process is a circuit that precisely captures the input-output behavior of the machine
learning system, which can then be used to reason about its behavior.
This includes explaining decisions, measuring robustness and formally proving properties. 

\begin{figure}[h]
  \centering
  \includegraphics[width=\linewidth]{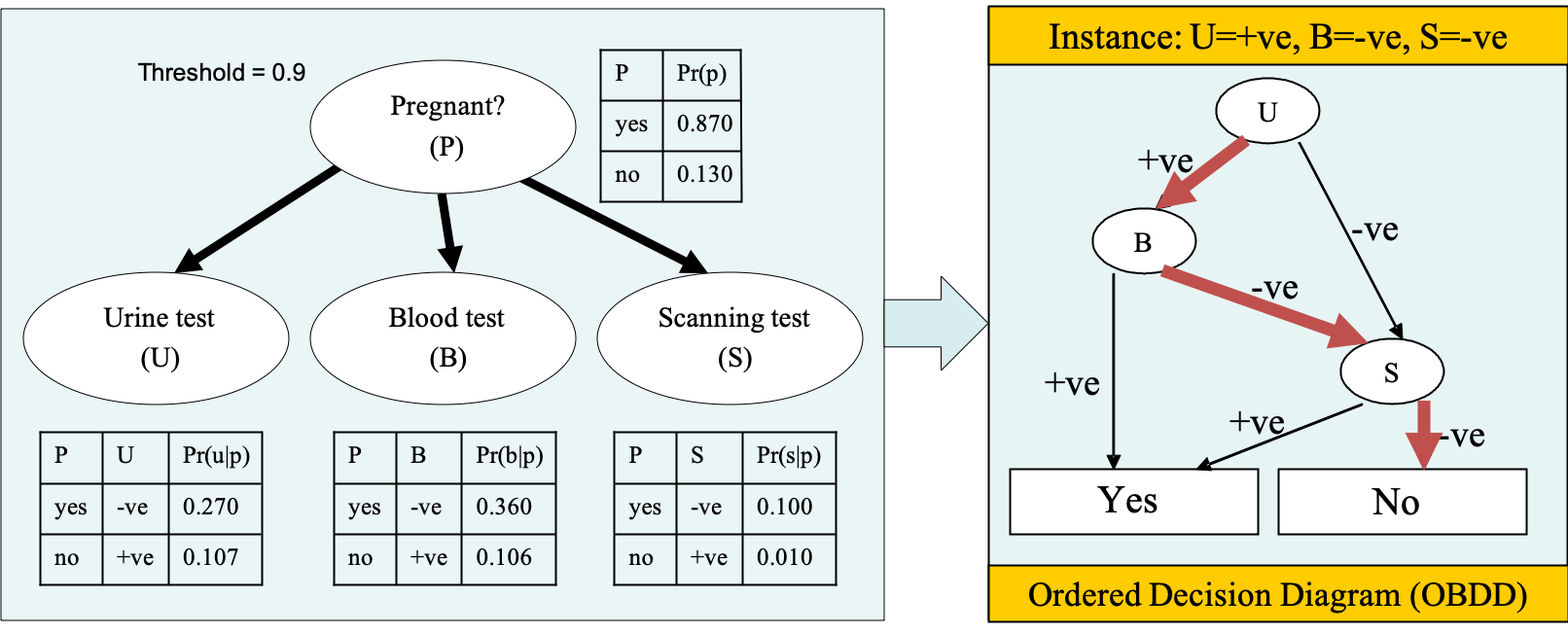}
  \caption{Compiling a Naive Bayes classifier into a symbolic decision graph (a tractable NNF circuit). \label{fig:nb2obdd} \label{fig:obdd}}
    \Description{Compiling machine learning systems into circuits.}
\end{figure}

Consider the example in Figure~\ref{fig:nb2obdd} which pertains to one of the simplest machine learning systems: 
a Naive Bayes classifier. We have class variable \(P\) and three features \(B\), \(U\) and \(S\). 
Given an instance (patient) and their test results \(b\), \(u\) and \(s\), this classifier renders a decision
by computing the posterior probability \(\pr(p\vert b,u,s)\) and then checking whether it passes a 
given threshold \(T\). If it does, we declare a positive decision; otherwise, a negative decision. 

While this classifier is numeric and its decisions are based on probabilistic reasoning, it does induce a 
discrete decision function. In fact, the function is Boolean in this case as it maps the Boolean variables \(B\), \(U\) and \(S\),
which correspond to test results, into a binary decision. This observation was originally 
made in~\cite{chanUAI03}, which proposed the compilation of Naive Bayes classifiers into symbolic decision
graphs as shown in Figure~\ref{fig:nb2obdd}. For every instance, the decision made by the (probabilistic)
naive Bayes classifier is guaranteed to be the same as the one made by the (symbolic) decision graph.

The compilation algorithm of~\cite{chanUAI03} generates Ordered Decision Diagrams (ODDs), 
which correspond to Ordered Binary Decision Diagrams (OBDDs) 
when the features are binary. Recall from Figure~\ref{fig:obdd2nnf}
that an OBDD corresponds to a tractable NNF circuit (once one adjusts for notation). Hence, the
proposal in~\cite{chanUAI03} amounted to compiling a Naive Bayes classifier into a tractable
NNF circuit that precisely captures its input-output behavior. 
This compilation algorithm was recently extended to Bayesian network classifiers with tree
structures~\cite{ShihCD18} and later to Bayesian network classifiers with arbitrary 
structures~\cite{ShihCD19}.\footnote{See \url{http://reasoning.cs.ucla.edu/xai/} for related software.}
Certain classes of neural networks can also be compiled into tractable circuits, which include SDD circuits as 
shown in~\cite{ChoiShiShihDarwiche18,SDC19,ShiShihDarwicheChoi20}.

While Bayesian and neural networks are numeric in nature, random forests
are not (at least the ones with majority voting). Hence, random forests represent less of a challenge
for this role of logic as we can easily encode the input-output behavior of a random forest
using a Boolean formula. We first encode each decision tree into a Boolean formula, which
is straightforward even in the presence of continuous variables (the learning algorithm discretizes
the variables). We then combine these formulas using a majority circuit. The remaining
challenge is purely computational as we now need to compile the Boolean formula into a
suitable tractable circuit.

We next turn to reasoning about the behavior of classifiers, assuming they have been
compiled into tractable circuits.

\subsection{Explaining Decisions}

Consider the classifier in Figure~\ref{fig:nb2obdd} and Susan who tested positive for the 
blood, urine and scanning tests. The classifier says that Susan is pregnant and we need to know why.

The first notion to address this question is the  {\em PI-explanation} introduced 
in~\cite{ShihCD18} and termed {\em sufficient reason} in~\cite{DarwicheHirth20a} (to make distinctions
with other types of reasons). A sufficient reason is a minimal set of instance characteristics that
is guaranteed to trigger the decision, regardless of what the other characteristics might be.
In this example, Susan would be classified as pregnant as long as she tests positive for the scanning test;
that is, regardless of what the other two test results are. Hence, \(S\!=\!\pos\) is a sufficient
reason for the decision. There is only one other sufficient reason for this 
decision: \(B\!=\!\pos,\:U\!=\!\pos\). Combining the two sufficient reasons we get 
 \(S\!=\!\pos \vee (B\!=\!\pos,\:U\!=\!\pos)\), which is called the {\em complete reason} behind
the decision~\cite{DarwicheHirth20a} or simply {\em the decision's reason.}

The reason behind a decision provides the most general abstraction of an instance that
can trigger the decision. Any instance property that can trigger the decision is captured by the
reason. The reason behind a decision can also be used to decide whether the decision is biased,
and in some cases whether the classifier itself is biased even when the considered decision
is not. We will provide concrete examples later but we first need
to establish the semantics of sufficient and complete reasons.

\begin{figure}[h]
  \centering
  \includegraphics[width=\linewidth]{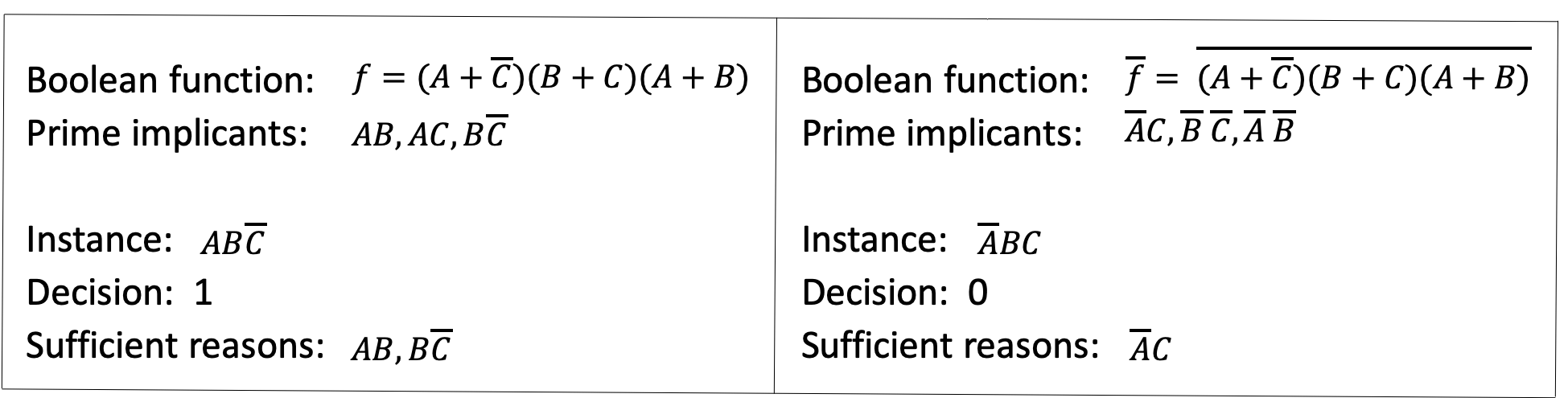}
  \caption{Prime implicants of Boolean functions. \label{fig:PI}}
    \Description{Prime implicants.}
\end{figure}

These notions are based on {\em prime implicants} of Boolean functions, which have been
studied extensively in the literature~\cite{BooleanFunctions,quine1,mccluskey,quine2}.
Consider the Boolean function \(f\) in Figure~\ref{fig:PI} over variables \(A\), \(B\) and \(C\).
A prime implicant of the function is a minimal setting of its variables that causes the
function to trigger. This function has three prime implicants as shown in the figure: \(AB\), \(AC\) and \(B\overline{C}\).
Consider now the instance \(AB\overline{C}\) leading to a positive decision \(f(AB\overline{C})=1\). 
The sufficient reasons for this decision are the prime implicants of function \(f\) that are 
compatible with the instance: \(AB\) and \(B\overline{C}\).
Explaining negative decisions requires working with the function's complement \(\overline{f}\).
Consider instance \(\overline{A}BC\), which sets the function to~\(0\). The complement \(\overline{f}\)
has three prime implicants \(\overline{A}C\), \(\overline{B}\,\overline{C}\) and \(\overline{A}\,\overline{B}\).
Only one of these is compatible with the instance, \(\overline{A}C\), so it is the
only sufficient reason for the decision on this instance.\footnote{The popular 
Anchor~\cite{ANCHOR} system can be viewed as computing approximations of sufficient reasons. 
The quality of these approximations has been evaluated on some datasets and corresponding 
classifiers in~\cite{JoaoApp}, where an approximation is called {\em optimistic} if it is a strict subset 
of a sufficient reason and  {\em pessimistic} if it is a strict superset of a sufficient reason.
Anchor computes approximate explanations without having to abstract the
machine learning system into a symbolic representation. Another set of approaches abstract
the behavior into symbolic form and compute sufficient reasons or other verification queries exactly,
but using SAT-based techniques instead of compiling into tractable circuits; see, e.g.,~\cite{KatzBDJK17,Leofante18,NarodytskaKRSW18,ShihCD18,IgnatievNM19a,IgnatievNM19b,SDC19}.}

\begin{figure*}[h]
  \centering
  \includegraphics[width=.16\linewidth]{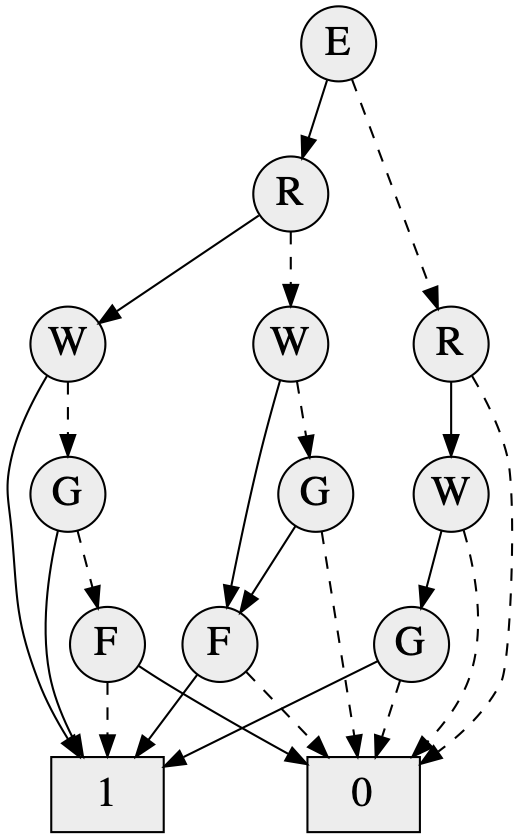}\hspace{5mm}
  \includegraphics[width=.38\linewidth]{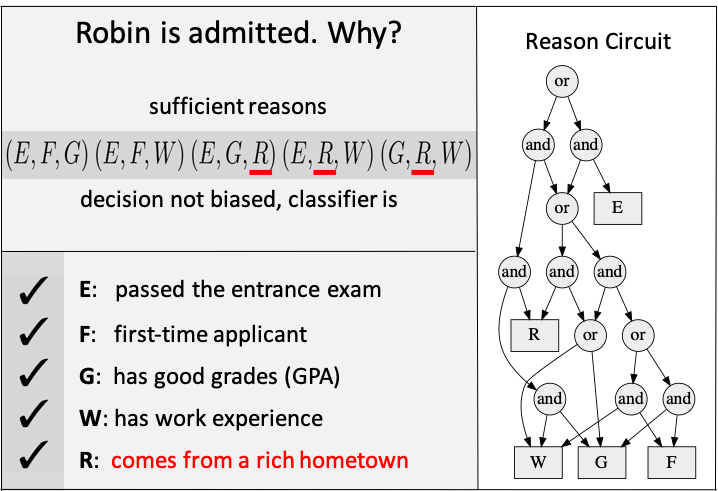}\hspace{5mm}
  \includegraphics[width=.38\linewidth]{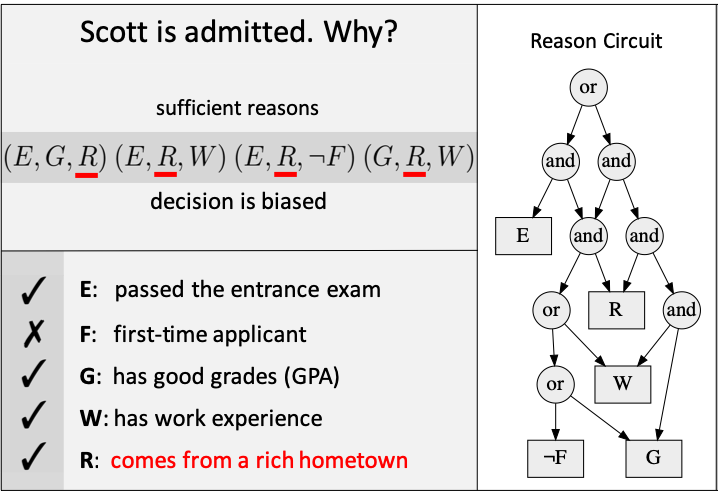}
  \caption{Explaining admission decisions of an OBDD classifier (left). The reason circuit
  represents the complete reason behind a decision (a disjunction of all its sufficient
  reasons). The reason circuit is monotone and hence tractable. \label{fig:adm}}
    \Description{Explaining decisions of machine learning systems.}
\end{figure*}

Sufficient reasons can be used to reason about decision and classifier bias, which are defined
in the context of {\em protected features.} A decision on an instance is biased iff it would be
different had we only changed protected features in the instance. A classifier is
biased iff it makes at least one biased decision. If every sufficient reason of a decision contains
at least one protected feature, the decision is guaranteed to be biased~\cite{DarwicheHirth20a}.
If some but not all sufficient reasons contain protected features, the decision is not biased
but the classifier is guaranteed to be biased (that is, the classifier will make a biased decision 
on some other instance).

Consider Figure~\ref{fig:adm} which depicts an admissions classifier in the form of an OBDD
(the classifier could have been compiled from a Bayesian network, a neural network or a random forest).
The classifier has five features, one of them is protected: whether the applicant comes from a rich
hometown (\(R\)). Robin is admitted by the classifier and the decision has five sufficient 
reasons, depicted in Figure~\ref{fig:adm}. Three of these sufficient reasons contain the protected feature \(R\)
and two do not. Hence, the decision on Robin is not biased, but the classifier is biased.
Consider now Scott who is also admitted. The decision on Scott has four sufficient reasons
and all of them contain a protected feature. Hence, the decision on Scott is biased:
it will be reversed if Scott were not to come from a rich hometown.

A decision may have an exponential number of reasons, which makes it impractical to analyze
decisions by enumerating sufficient reasons. One can use the complete reason behind
a decision for this purpose as it contains all the needed information. Moreover, if the classifier
is represented by an appropriate tractable circuit, then the complete reason behind a decision
can be extracted from the classifier in linear time, in the form of another tractable circuit 
called the {\em reason circuit}~\cite{DarwicheHirth20a}. Figure~\ref{fig:adm} depicts the
reason circuits for decisions on Robin and Scott. Reason circuits get their tractability from being
monotone, allowing one to efficiently reason about decisions including their bias.
One can also reason about counterfactuals once the reason circuit for a decision is
constructed. For example, one can efficiently evaluate statements such as:
The decision on April would stick {\em even if} she were not to have work experience 
{\em because} she passed the entrance exam; see~\cite{DarwicheHirth20a} for details.

\begin{figure}[tb]
  \centering
  \includegraphics[width=.33\linewidth]{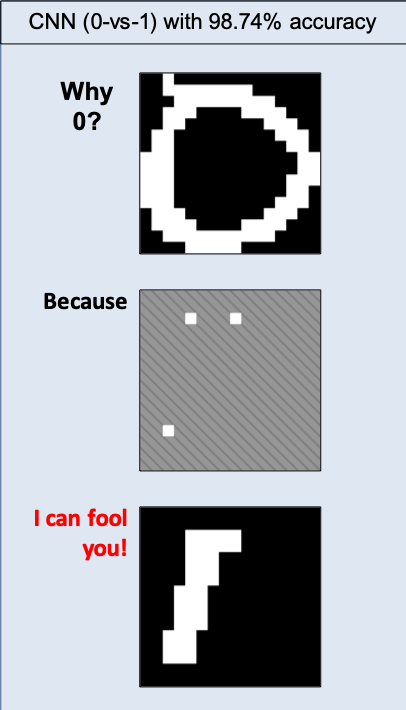}
  \caption{Explaining the decisions of a neural network. \label{fig:digits}}
    \Description{Explaining decisions of neural networks.}
\end{figure}

We conclude this section by pointing to an example from~\cite{ShiShihDarwicheChoi20},
which compiled a Convolutional Neural Network (CNN) that classifies digits \(0\)
and \(1\) in \(16 \times 16\) images. The CNN had an accuracy of \(98.74\%\). Figure~\ref{fig:digits}
depicts an image that was classified correctly as containing digit \(0\). One of the sufficient
reasons for this decision is also shown in the figure, which includes only \(3\) pixels out of \(256\).
If these three pixels are kept white, the CNN will classify the image as containing digit \(0\) regardless
of the state of other pixels. 

\subsection{Robustness and Formal Properties}

One can define both decision and model robustness. The robustness of a decision is
defined as the smallest number of features that need to flip
before the decision flips~\cite{ShihCD18b}. Model robustness is defined
as the average decision robustness (over all possible instances)~\cite{ShiShihDarwicheChoi20}. 
Decision robustness is $\coNP$-complete and model robustness is $\sp$-hard.
If the decision function is represented using a tractable circuit of a suitable type, which
includes OBDDs, then decision robustness can be computed in time linear in the
circuit size~\cite{ShihCD18b}. Model robustness can be computed using a sequence
of polytime operations but the total complexity is not guaranteed to be in polytime~\cite{ShiShihDarwicheChoi20}.

\begin{figure}[tb]
\centering
  \includegraphics[width=.75\linewidth]{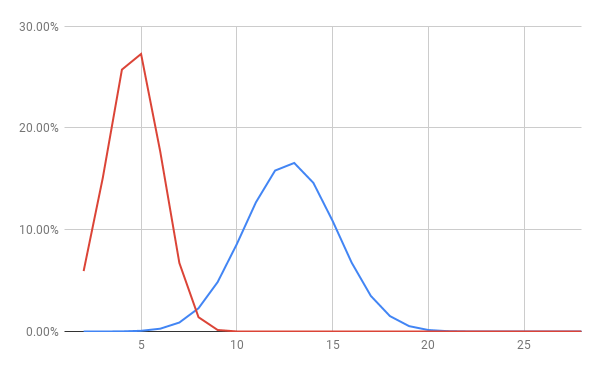}
  \label{fig:model_robustness}
\caption{Robustness level vs. proportion of instances for two neural networks with similar accuracies.
Net~$1$ is plotted in blue (right) and Net~$2$ in red (left).
\label{fig:robust}}
  \Description{Robustness of neural networks.}
\end{figure}

Figure~\ref{fig:robust} depicts an example of robustness analysis for two CNNs that classify
digits \(1\) and \(2\) in \(16 \times 16\) images~\cite{ShiShihDarwicheChoi20}.
The two CNNs have the same architectures but were trained using two different parameter seeds, leading
to testing accuracies of $98.18$ (Net~$1$) and $96.93$ (Net~$2$). The CNNs were compiled into SDD circuits
where the SDD of Net~$1$ had $3,653$ edges and  the one for Net~$2$ had only $440$ edges. 
The two CNNs are similar in terms of accuracy (differing by only $1.25\%$) but are very different when 
compared by robustness. For example, Net~$1$ attained a model robustness of $11.77$ but Net~$2$ 
obtained a robustness of only $3.62.$ For Net~$2$, this means that on average, $3.62$ pixel flips are needed to flip a digit-$1$ 
classification to digit-$2$, or vice versa. Moreover, the maximum robustness of Net~$1$ was $27$, while that of Net~$2$ 
was only $13.$ For Net~$1$, this means that there is an instance that would not flip classification unless one
is allowed to flip at least $27$ of its pixels. 
For Net~$2$, it means the decision on any instance can be flipped if one is allowed to flip $13$ or more pixels.
Note that Figure~\ref{fig:robust} reports the robustness of \(2^{256}\) instances for each CNN, which
is made possible by having captured the input-output behavior of these CNNs using tractable circuits.

In the process of compiling a neural network into a tractable circuit as proposed in~\cite{ChoiShiShihDarwiche18,ShiShihDarwicheChoi20},
one also compiles each neuron into its own tractable circuit. This allows one
to interpret the functionality of each neuron by analyzing the corresponding tractable circuit, which is a function
of the network's inputs. For example, if the tractable circuit supports model counting in polytime, then one can efficiently answer
questions such as: Of all network inputs that cause a neuron to fire, what proportion of them set input \(X_i\) to $1$?
This is just one mode of analysis that can be performed efficiently once the input-output behavior of a machine
learning system is compiled into a suitable tractable circuit. Other examples include monotonicity analysis, which is
discussed in~\cite{ShihCD18b}.

\shrink{
\begin{figure}[tb]
\centering
\includegraphics[width=.3\linewidth]{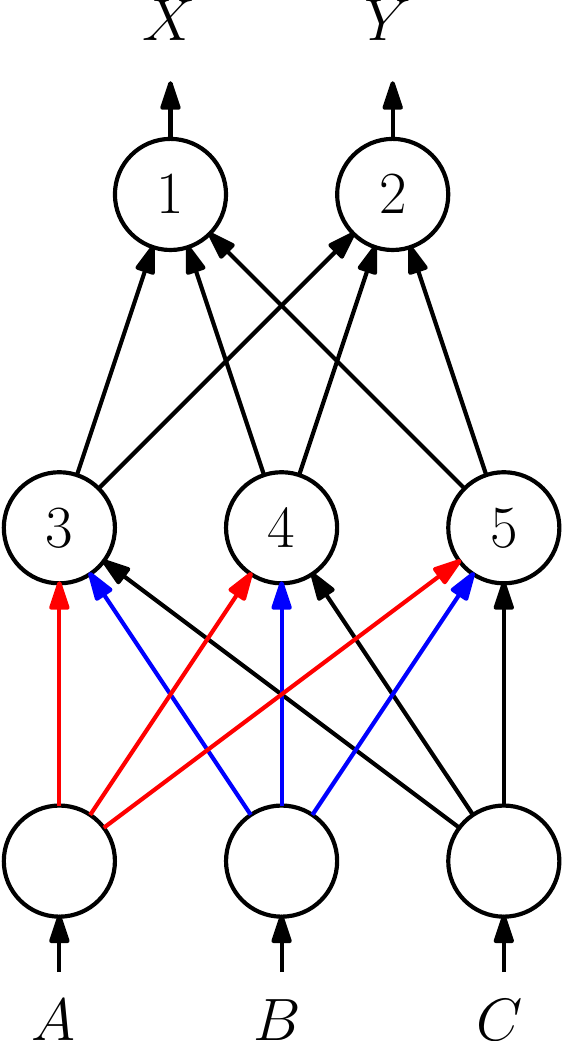}\hspace{3mm}
\includegraphics[width=.3\linewidth]{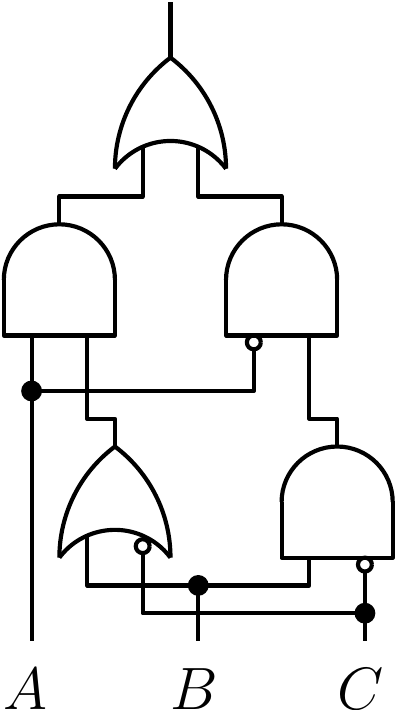} \hspace{3mm}
\includegraphics[width=.3\linewidth]{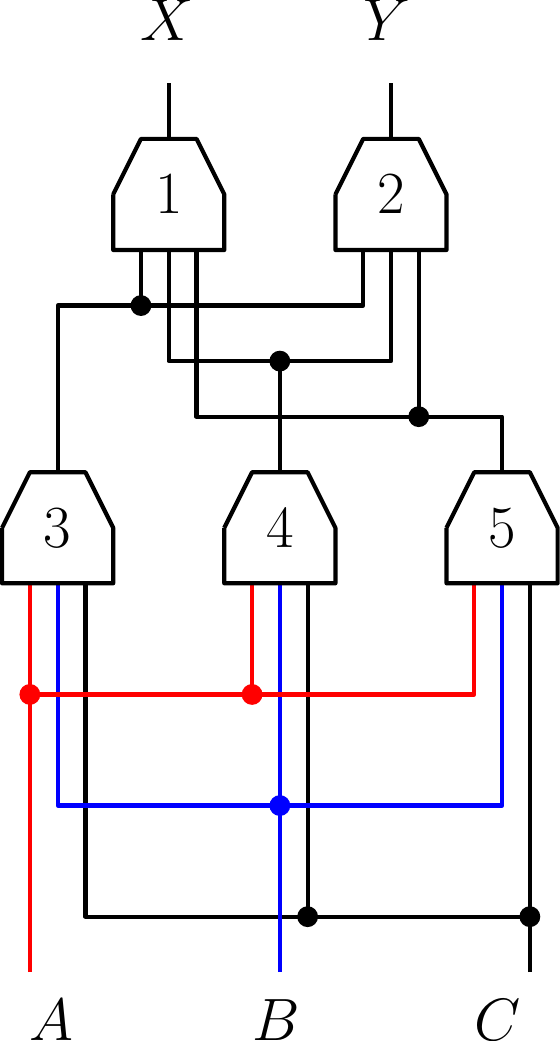}
\caption{A neural network, the circuit of a single neuron, and the circuit of the original network.  \label{fig:nn-compile}}
  \Description{}
\end{figure}
}

\section{Conclusion and Outlook}
\label{sec:outlook}

We reviewed three modern roles for logic in artificial intelligence: logic as a basis for computation,
logic for learning from a combination of data and knowledge, and logic for reasoning about the 
behavior of machine learning systems.

Everything we discussed was contained within the realm of propositional logic and based on 
the theory of tractable Boolean circuits. The essence of these roles is based on an ability to 
compile Boolean formula into Boolean circuits with appropriate properties. 
As such, the bottleneck for advancing these roles further---at least as far as enabling more 
scalable applications---boils down to the need for improving the effectiveness of knowledge 
compilers (i.e., algorithms that enforce certain properties on NNF circuits). 
The current work on identifying (more) tractable circuits and further studying their properties 
appears to be ahead of work on improving compilation algorithms. 
This imbalance needs to be addressed, perhaps by encouraging more open source
releases of knowledge compilers and the nurturing of regular competitions as has been
done successfully by the SAT community.\footnote{\url{www.satlive.org}}

McCarthy's proposal for using logic as the basis for knowledge representation and
reasoning had an overwhelming impact~\cite{McCarthy59}. But it also entrenched
into our thinking a role for logic that has now been surpassed, while continuing to be
the dominant role associated with logic in textbooks on artificial intelligence. 
Given the brittleness of purely symbolic representations in key applications,
and the dominance today of learned, numeric representations, this entrenched association
has pushed logic to a lower priority in artificial intelligence research than what can 
be rationally justified.  
  
Another major (and more recent) transition relates to the emergence of machine
learning ``boxes,'' which have expanded the scope and utility of symbolic 
representations and reasoning. While symbolic representations may be too brittle
to fully represent certain aspects of the real world, they are provably sufficient
for representing the behavior of certain machine learning boxes (and exactly when
the box's inputs/outputs are discrete). This has created a new role for logic 
in ``reasoning about what was learned'' in contrast to the older and entrenched 
role of reasoning about the real world. 

The latter transition is more significant than may appear on first sight. For example,
the behavior of a machine learning box is driven by the box's internal causal mechanisms
and can therefore be subjected to causal reasoning and analysis---even when the
box itself was built using ad hoc techniques and may have therefore missed on capturing
causality of the real world. Machine learning boxes, called ``The AI'' by many today, 
are now additional inhabitants of the real world. As such, they should be viewed 
as a new ``subject'' of logical and causal reasoning perhaps more so than their competitor. 

These modern transitions---and the implied new modes of interplay between logic,
probabilistic reasoning and machine learning---need to be paralleled by a
transition in AI education that breaks away from the older models of viewing
these areas as being either in competition or merely in modular harmony. 
What we need here is not only {\em integration} of these methods but also their {\em fusion.}
Quoting~\cite{DBLP:journals/cacm/Darwiche18}: ``We need a new generation of AI 
researchers who are well versed in and appreciate classical AI, machine learning, 
and computer science more broadly while also being informed about AI history.''
Cultivating such a generation of AI researchers
is another bottleneck to further advance the modern roles of logic  in AI
and to further advance the field of AI as a whole. 

%
\begin{acks}
I wish to thank Arthur Choi and Jason Shen for their valuable feedback and help with some of the figures.
This work has been partially supported by grants from NSF IIS-1910317, 
ONR N00014-18-1-2561, DARPA N66001-17-2-4032 and a gift from JP Morgan.
\end{acks}

%
\bibliographystyle{ACM-Reference-Format}
\bibliography{bib/kr16,bib/kr20,bib/ecai20,bib/aaai19,bib/aaai18}

\end{document}